\documentclass[a4paper,fleqn]{cas-dc}

\usepackage[authoryear]{natbib}
\usepackage{url}

\usepackage{breakurl}
\usepackage{soul}

\usepackage[section]{placeins}

\def\tsc#1{\csdef{#1}{\textsc{\lowercase{#1}}\xspace}}
\tsc{WGM}
\tsc{QE}
\tsc{EP}
\tsc{PMS}
\tsc{BEC}
\tsc{DE}
\newcommand{\Fone}{F\textsubscript{1}}
\begin{document}
\let\WriteBookmarks\relax
\def\floatpagepagefraction{1}
\def\textpagefraction{.001}
\shorttitle{Flood severity mapping from Volunteered Geographic Information by interpreting water level from images containing people}
\shortauthors{Yu Feng et~al.}

\title [mode = title]{Flood severity mapping from Volunteered Geographic Information by interpreting water level from images containing people: a case study of Hurricane Harvey}
%\tnotemark[1,2]
%\tnotetext[1]{This document is the results of the research project funded by the National Science Foundation.}
%\tnotetext[2]{The second title footnote which is a longer text matter to fill through the whole text width and overflow into another line in the footnotes area of the first page.}

\author[1]{Yu Feng}%[orcid=0000-0001-5110-5564]
\cormark[1]
\ead{yu.feng@ikg.uni-hannover.de}

\author[1]{Claus Brenner}
%\ead{claus.brenner@ikg.uni-hannover.de}

\author[1]{Monika Sester}
%\ead{monika.sester@ikg.uni-hannover.de}

\address[1]{Institute of Cartography and Geoinformatics, Leibniz University Hannover, Appelstra\ss{}e 9a,
30167 Hannover, Germany}

\cortext[cor1]{Corresponding author}

%\author[1,3]{CV Radhakrishnan}[type=editor,
%                        auid=000,bioid=1,
%                        prefix=Sir,
%                        role=Researcher,
%                        orcid=0000-0001-7511-2910]
%\cormark[1]
%\fnmark[1]
%\ead{cvr_1@tug.org.in}
%\ead[url]{www.cvr.cc, cvr@sayahna.org}
%
%\credit{Conceptualization of this study, Methodology, Software}
%
%\author[2,4]{Han Theh Thanh}[style=chinese]
%
%\author[2,3]{CV Rajagopal}[%
%   role=Co-ordinator,
%   suffix=Jr,
%   ]
%\fnmark[2]
%\ead{cvr3@sayahna.org}
%\ead[URL]{www.sayahna.org}
%
%\credit{Data curation, Writing - Original draft preparation}
%
%\address[2]{Sayahna Foundation, Jagathy, Trivandrum 695014, India}
%
%\author%
%[1,3]
%{Rishi T.}
%\cormark[2]
%\fnmark[1,3]
%\ead{rishi@stmdocs.in}
%\ead[URL]{www.stmdocs.in}
%
%\address[3]{STM Document Engineering Pvt Ltd., Mepukada, Malayinkil, Trivandrum 695571, India}
%
%\cortext[cor1]{Corresponding author}
%\cortext[cor2]{Principal corresponding author}
%\fntext[fn1]{This is the first author footnote. but is common to third author as well.}
%\fntext[fn2]{Another author footnote, this is a very long footnote and it should be a really long footnote. But this footnote is not yet sufficiently long enough to make two lines of footnote text.}
%
%\nonumnote{This note has no numbers. In this work we demonstrate $a_b$ the formation Y\_1 of a new type of polariton on the interface between a cuprous oxide slab and a polystyrene micro-sphere placed on the slab.}

\begin{abstract}
With increasing urbanization, in recent years there has been a growing interest and need in monitoring and analyzing urban flood events. Social media, as a new data source, can provide real-time information for flood monitoring. The social media posts with locations are often referred to as Volunteered Geographic Information (VGI), which can reveal the spatial pattern of such events. Since more images are shared on social media than ever before, recent research focused on the extraction of flood-related posts by analyzing images in addition to texts. Apart from merely classifying posts as flood relevant or not, more detailed information, e.g. the flood severity, can also be extracted based on image interpretation. However, it has been less tackled and has not yet been applied for flood severity mapping.

In this paper, we propose a novel three-step process to extract and map flood severity information. First, flood relevant images are retrieved with the help of pre-trained convolutional neural networks as feature extractors. Second, the images containing people are further classified into four severity levels by observing the relationship between body parts and their partial inundation, i.e. images are classified according to the water level with respect to different body parts, namely ankle, knee, hip, and chest. Lastly, locations of the Tweets are used for generating a map of estimated flood extent and severity. This process was applied to an image dataset collected during Hurricane Harvey in 2017, as a proof of concept. The results show that VGI can be used as a supplement to remote sensing observations for flood extent mapping and is beneficial, especially for urban areas, where the infrastructure is often occluding water. Based on the extracted water level information, an integrated overview of flood severity can be provided for the early stages of emergency response.
\end{abstract}

%\begin{abstract}
%This template helps you to create a properly formatted \LaTeX\ manuscript.
%\noindent\texttt{\textbackslash begin{abstract}} \dots 
%\texttt{\textbackslash end{abstract}} and
%\verb+\begin{keyword}+ \verb+...+ \verb+\end{keyword}+ 
%which contain the abstract and keywords respectively. 
%\noindent Each keyword shall be separated by a \verb+\sep+ command.
%\end{abstract}

%\begin{graphicalabstract}
%\includegraphics{figs/grabs.pdf}
%\end{graphicalabstract}

%\begin{highlights}
%\item Research highlights item 1
%\item Research highlights item 2
%\item Research highlights item 3
%\end{highlights}


\maketitle

%\tableofcontents

\section{Introduction}

%% KAO: Sloppy spacing ensures non-overfull lines. Can be removed if this is not an issue.
\sloppy

Flood, as one of the great natural disasters, endangers people's safety and their property. 
In the last few decades, the density of urban development and the area of sealed land has increased, which leads to more severe flooding situations than ever before \citep{maard2018urbanization}. 
%\citep{maard2018urbanization}. 
Intensive studies have been conducted on flood extent mapping from satellite remote sensing data. Methods for flood detection have been tested on different high-resolution remote sensing products, such as Landsat TM/ETM+
%Thematic Mapper~/ Enhanced Thematic Mapper Plus (TM/ETM+) 
\citep{li2015sub}, MODIS 
%(Moderate Resolution Imaging Spectroradiometer) 
\citep{son2013satellite}, and TerraSAR-X \citep{martinis2015fully,li2019urban}. %\cite{malinowski2016local} utilized airborne LiDAR for flood extent mapping. 
Researchers used the  Normalized Difference Water Index (NDWI) \citep{huang2018reconstructing}, modified NDWI \citep{rosser2017rapid} or image semantic segmentation \citep{sarker2019flood} to obtain the water extent. With a given Digital Terrain Model (DTM), water depth can be further estimated \citep{singh2015evaluation}. 

However, %\hl{optical??} 
airborne or satellite remote sensing products can hardly achieve a real-time monitoring of flood events for the following three reasons. First, severe weather conditions limit the visibility of both products, especially because of the clouds along with heavy rain \citep{huang2018reconstructing}. Second, the revisit time of the satellites limits the data availability \citep{feng2015urban}. Commercial optical satellites sometimes need several days after an event to acquire high-resolution imagery \citep{ning2020prototyping}. Third, airborne sensors such as Unmanned Aerial Systems (UAS) normally can only be deployed with a controllable risk after the events. All these limitations may result in the loss of first-hand information on a flood event.
For the flood events in urban areas, especially floods caused by short-time storm and heavy rainfall, observations from remote sensing are not able to achieve a satisfactory spatial and temporal resolution. Therefore, observations from the ground are needed as a supplement to traditional earth observation methods.

Based on the ``citizens as sensors'' idea, crowdsourcing has been identified as a new approach to gather geospatial data \citep{heipke2010crowdsourcing}. Volunteered Geographic Information (VGI) \citep{goodchild2007citizens} are the crowdsourced geospatial data, which could be map elements (e.g. OpenStreetMap), user uploaded GPS trajectories, and also geotagged user-generated texts and photos. Crowdsourcing can be conducted with two approaches, participatory or opportunistic. The participatory approach requires active participation by the users. For example, mobile apps have been developed to provide citizens with a platform to report for desired disaster events, such as ``Did You Feel It?'' from USGS for earthquake crowdsourcing \citep{atkinson2007did}. Nevertheless, motivating users to participate and provide information is difficult. As stated in the 90:9:1 rule observed by \cite{nielsen2016rule} for social media and online communities, only 1\% of the users participate frequently and are responsible for most contributions while 90\% only use or read. The remaining 9\% contribute from time to time. Therefore, an opportunistic approach is desirable, where the information is acquired in a quasi-unconscious and passive manner, for instance, from social media.

\begin{figure*}
	\centering
	\includegraphics[clip=true,trim=60pt 260pt 100pt 40pt, width=1.8\columnwidth]{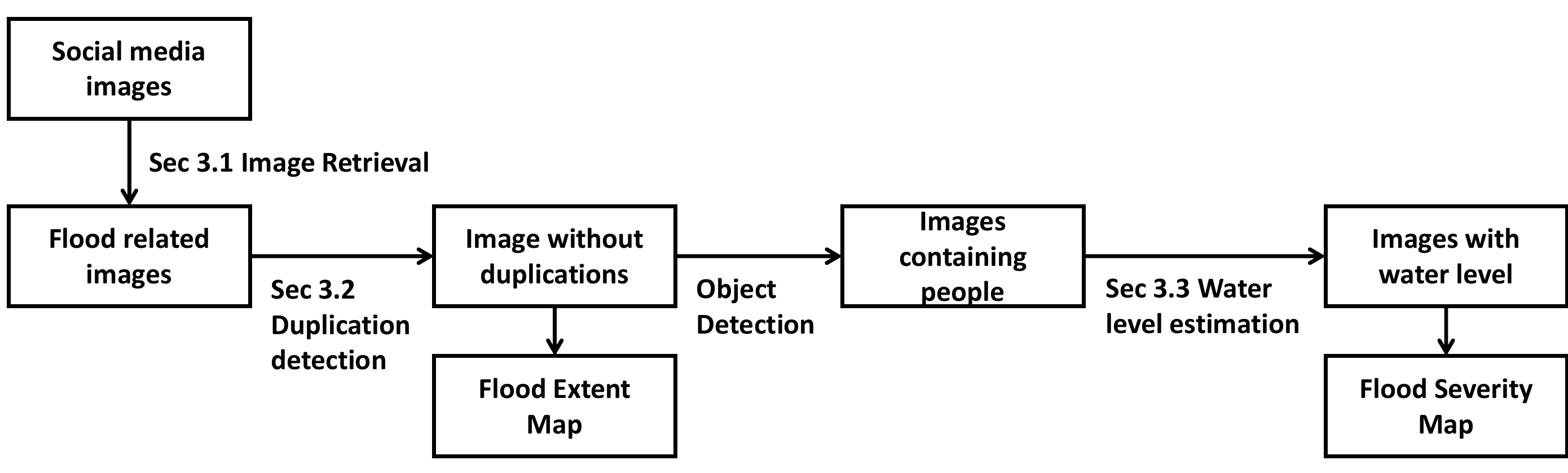}
	\caption{Workflow of our proposed process to extract flood extent and flood severity from social media data}
	\label{fig:pipeline}
\end{figure*}

Social media offers the possibility to collect thematic, spatio-temporal information in real time. It is nowadays frequently used in emergency response. The emergency services such as 911 are often overloaded when a crisis happens, and people in the affected area often seek for help from social media~\citep{cowan2017when}. In this case, the social media act as a platform, where critical information can be shared \cite[e.g. Facebook Crisis Response, ][]{iyengar2015facebook}. Even though flood relevant information occupies only a very small proportion of the social media data streams, the geotagged flood relevant posts can still contribute to flood monitoring and extent mapping \citep{huang2018reconstructing}. Such real-time information can improve the situation awareness of people in the flooding zones. It is also an essential information source for the city managers at the response stage of the disaster.

\cite{assumpccao2018citizen} summarized that flood-related information such as water level, velocity, and flood extent can be extracted from social media and used for flood monitoring, mapping and modelling purposes. Almost all of the research summarized in their review needs a human annotator to interpret the information from social media texts and images. In order to automate this process, flood-related posts can be retrieved from social media using convolutional neural networks \cite[e.g.][]{feng2018extraction, huang2018visual}. They are used for monitoring and mapping of the flood extent.
\cite{le2016crowdsourced} estimate the water surface velocity from YouTube videos. Nevertheless, the automatic extraction of \textit{water level} information from images has rarely been tackled. The water depth is normally determined through human inspection by comparison with objects in the images that have approximately known dimensions \citep{assumpccao2018citizen}, e.g. people standing in the water or wheels of cars in the water \citep{kutija2014model}. 

This interpretation is relatively easy for humans, however, it is a nontrivial problem for computers. Even though modern deep learning technologies can successfully interpret the relevance of the photos or texts to flooding events, the extraction of more detailed severity information from images has been presented in only a few papers \citep[e.g.][]{chaudhary2019flood, pereira2019assessing}. 
\cite{pereira2019assessing} classified images into three categories (i.e. no flood, below 1 m and above 1 m) based on global deep features and suggested to achieve fine-grained water level estimation according to partially submerged objects with approximate known dimensions in their future work. 
\cite{chaudhary2019flood} estimated the water level via local deep features around objects with approximate known dimensions (i.e. person, car, bus, bicycle, and house), however, required a time-consuming pixel-level annotation.
Nevertheless, none of these approaches has further used the extracted information for flood severity mapping.

Therefore, this paper aims to extract fine-grained flood severity information from social media images with less annotation effort (i.e. one label per image) by using component level information (i.e. human pose). 
We target on images containing people in flood scenarios. By analyzing the relationship between body parts and water, we can obtain an estimation of the water level in the scenarios. We aim to investigate whether such component level information is beneficial by comparing with the baselines, which use either global deep feature of the whole image or local deep features around detected people.

In further, we aim to investigate the usefulness of such extracted information for flood extent and severity mapping. Each image predicting a water level can be associated with the time and location provided by users' mobile devices. Subsequently, a map can be generated with post locations and water level estimations. In this way, the large collection of individual, local information is aggregated to generate maps of flood extent and severity.

The remainder of this paper is organized as follows: Section \ref{sec:related_works} introduces related work. In Section \ref{sec:methodology_experiments}, the methods used for retrieval of flood relevant social media posts and classification of the images into four severity levels are introduced, together with the experiments. As a proof of concept for the proposed process, Section \ref{sec:case_study} presents a study for the case when Hurricane Harvey raged Texas in the United States in 2017, showing the extent and severity mapping results from social media posts. Furthermore, we compared the extracted extent and severity information with existing mapping results from official sources. 
A conclusion and an outlook are given in the last section.
The overview of our whole proposed workflow is visualized in Figure~\ref{fig:pipeline}, where the main components and their corresponding section numbers are presented.

\section{Related work}
\label{sec:related_works}

In the last few years, many studies have been conducted to extract water level information from citizen observations. For instance, apps and websites were developed to provide citizens with a platform to send water level gauge readings \citep[e.g.][]{lowry2013crowdhydrology, degrossi2014flood}. This quantitative information can be used directly without further analysis. 

More often, texts and images are collected from citizens via apps or social media. Researchers manually analyze them to extract qualitative observations about the water level.
Text information is mainly used for the filtering of relevant posts \citep[e.g.][]{fohringer2015social}. A few studies \citep[e.g.][]{eilander2016harvesting, smith2017assessing, li2018novel} derived water level from user-generated texts. 
\cite{eilander2016harvesting} applied flood mapping based on 888 water level mentions from social media texts during three days in 2015 in Jakarta, Indonesia, which is the most user active city on Twitter. Combined with DEM and hydraulic models, flood extent and water depth maps were generated. During this event, project PetaJakarta \citep{ogie2019crowdsourced} sent invitation requests to users who mentioned flood related keywords to participate in a collaborate flood mapping initiative and provide water level information \citep{see2019review}. However, without such a participatory mechanism, indications of flood depth are often absent \citep{smith2017assessing}.

The majority of the water level estimations are based on images. For example, participatory approaches have been conducted to collect pictures about a flood event in Newcastle, UK, on the 28\textsuperscript{th} of June 2012 from the citizens \citep{kutija2014model}. 12 images for 12 different places were manually annotated with water depth and used as validation for flood models. 
In the project \textit{RiskScape} in Christchurch and Dunedin, New Zealand, people were asked to send photos of flood levels with time and place information after the flood peak. In Christchurch, 600 photos were received and assessed by professionals. However, the project in Dunedin was discontinued due to a lack of response \citep{le2016crowdsourced}. 

Continuous engagement of people for contribution is hard to achieve, especially when people do not directly benefit from such projects. 
Therefore, researchers focus more on opportunistic approaches and extract the flood-relevant information from social media. 
For the fluvial flood in 2013 in Dresden, Germany, Tweets were filtered by flood-related keywords. Experts or voluntary annotators were asked to estimate the relevance regarding inundation mapping and the water level from social media photos on a web-based platform. Five inundation depth estimates were used to improve flood extent mapping \citep{fohringer2015social}. 
Field visits are necessarily needed if the precise water level and exact locations are required for the post-flood mapping. Real-time kinematic GNSS and conventional survey methods were used to verify the exact locations. For instance, 23 selected places with photos from Flickr and Facebook regarding the flood in Brisbane, Australia in January 2011 were verified after the flood using this method \citep{mcdougall2012use}.
From the above review, it can be concluded that current approaches, which make use of social media, need the involvement of human annotators. However, manual checks of flood relevant information from social media images are time-consuming and field visits are expensive.

Keyword filtering is a popular approach of many early studies to extract flood relevant posts~\citep[e.g.][]{fuchs2013tracing,li2018novel}.
However, due to the ambiguity of the keywords, the precision of retrieval using keyword filtering is limited. Many Natural Language Processing (NLP) methods have also been applied for retrieval of the on-topic posts, such as applying classic machine learning methods on \textit{TFIDF} \citep{manning2008introduction} or \textit{word2vec} \citep{mikolov2013distributed} features \cite[e.g.][]{hanif2017flood, bischke2017detection, bai2015sina, feng2018extraction}. \cite{feng2018ensembled} used \textit{fasttext} features \citep{bojanowski2017enriching} and summarized the single sentences using CNN or LSTM based network structures for text classification. 
However, water level information rarely appears in the texts and is mostly subjective. 
In this work, we mainly focus on the visual information of social media posts.

With the rapid development of computer vision, automatic interpretation of the flood relevant images using deep convolutional neural networks (CNN) has been introduced. Our previous work~\citep{feng2018extraction} focused on training CNN based classifiers for text and images to retrieve flood relevant social media posts. The models were applied on the data collected during the urban flood events in Paris and London in 2016. Spatial and temporal pattern of the flood related geo-tagged Tweets were analyzed. \textit{Multimedia Satellite (MMSat) Task}~\citep{bischke2017multimedia} in the \textit{MediaEval'17} benchmarking initiative focused on the retrieval of the flood relevant Flickr posts. Many teams working on this task also applied CNN based classifiers for textual and visual information. 
In another work, similar models were also applied to classify Twitter Tweets during Hurricane Harvey in 2017~\citep{huang2018visual}. 
Additionally, the ensemble of features from multiple pre-trained CNN was proved to be beneficial in some of the research (e.g.~\citealt{ahmad2017cnn, ahmad2018comparative}).
Using a classification, a large amount of information irrelevant to flood can be eliminated. 
Besides classification, there are also studies focusing on image retrieval. \cite{barz2018enhancing} proposed an approach, which can not only retrieve flood relevant images, but also images containing evidence for an inundation depth estimation. However, experts are still needed to extract the desired severity information manually, such as water level, based on these retrieved photos.

Automatic interpretation of the water level from crowdsourcing images has been tackled in only a few papers.
\cite{pereira2019assessing} classified water severity into three classes, namely no flood, below 1~m, and above 1~m. They used the DenseNet \citep{huang2017densely} and EfficientNet \citep{tan2019efficientnet} neural network architectures, where only the global deep features of the whole images are considered.
\cite{chaudhary2019flood} extended the Mask R-CNN model \citep{he2017mask} for water level classification. Images are annotated pixel-wise, which is very time-consuming. Objects, such as person, car, bus, bicycle, and house, are considered for flood level estimation. Images are classified into 11 water levels. In this case, the local deep features around the objects are considered for predicting the water level. 
In a very recent work, \cite{quan2020flood} also made use of human pose and object detection to predict the water level but only in two categories (i.e. above the knee and below the knee). The method is specifically designed for this binary classification task, where pre-defined rules were applied by observing the relation between body keypoints and person segments.
Multiple empirical thresholds were applied on ratios between different body parts to represent such a water level situation.

In this paper, we propose a novel method in Section \ref{sec:wlevel} that can automatically provide an estimation of water depth based on analyzing images of persons standing in the water. Object detection, human keypoint detection, and semantic segmentation using pre-trained deep learning models provide the primary information for this classification. We furthermore investigate the performance of our model by comparing it with approaches using global deep features and local deep features as baselines. 

\section{Methodology and experiments of flood evidence detection and water level estimation}
\label{sec:methodology_experiments}
The approach proposed in this paper has three main components, namely (1)~retrieval of flood relevant social media posts, (2)~duplication detection and (3)~water level estimation from images containing persons.

\subsection{Retrieval of flood relevant social media images}
\label{sec:retrieval}
Social media covers various contents. In order to extract flood relevant VGI from massive social media data, a retrieval step is always essential for all kinds of further applications. 

\subsubsection{Method}
\label{sec:retrieval_method}

\begin{figure}
    \includegraphics[clip=true,trim=230pt 290pt 200pt 40pt, width=\columnwidth]{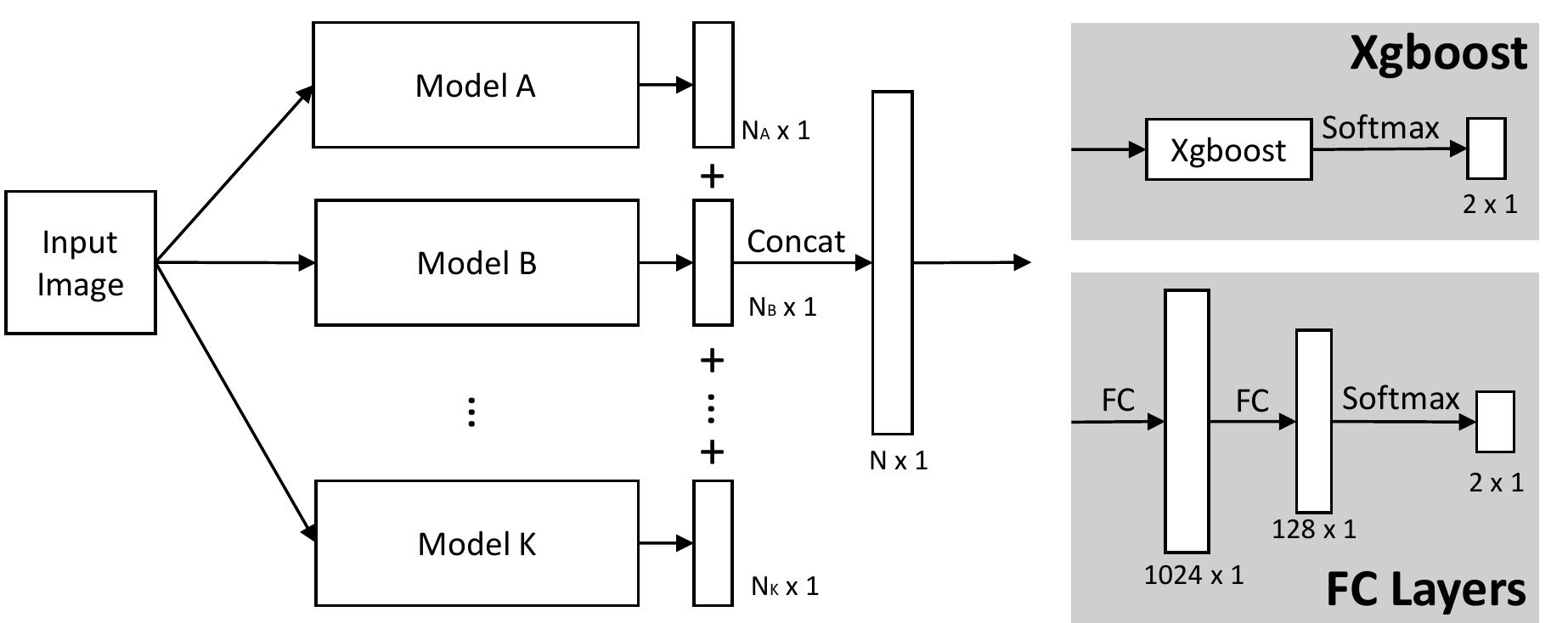}
    \caption{Late-fused model for image classification into two classes: flood relevant or not flood relevant}
    \label{fig:imageonly}
\end{figure}

Most of the current methods to extract visual features are based on the concept of transfer learning, where models pre-trained on ImageNet (e.g. in~\citealt{ahmad2017cnn, lopez2017multi, feng2018extraction, feng2018ensembled, huang2019linking}) or Place365 (e.g. in~\citealt{ahmad2017convolutional}) datasets are used instead of training a CNN model from scratch.
Therefore, we also used such a network architecture, similar to the one already used in \cite{feng2018ensembled}. 
We used features from a pre-trained single model or a concatenation of features from multiple pre-trained models. The images were classified  based on these features using either Xgboost \citep{chen2016xgboost} or fully-connected (FC) layers both with two softmax outputs (shown in Figure \ref{fig:imageonly}). The FC layers consist of two dense layers followed by batch normalization. Dropout of 50\% is applied at the output layer.
The softmax outputs on the positive class provide the confidence score of flood relevance. The final class prediction is based on a 0.5 threshold of this score.

Different pre-trained models are available, which were considered as the basic feature extractors: \textit{InceptionV3}~\citep{szegedy2016rethinking}, \textit{DenseNet201}~\citep{Huang_2017_CVPR}, \textit{InceptionResNetV2}~\citep{szegedy2017inception}. They were all trained based on ImageNet and could achieve a top-5 classification accuracy of 0.936, 0.937 and 0.953, respectively. Since CNN models pre-trained on Places365~\citep{zhou2017places} were reported to have a better performance due to their scene-level features~\citep{ahmad2019automatic}, a VGG16 network pre-trained on Places365~\citep{gkallia2017keras_places365} was considered in addition.

\subsubsection{Dataset}

\begin{table*}[width=1.9\linewidth,cols=5,pos=h]
\caption{Number of positive and negative examples for dataset}%\label{tbl1}
\begin{tabular*}{\tblwidth}{@{} LLL@{} }
\toprule
Dataset & Number of Neg. Examples & Number of Pos. Examples\\
\midrule
DIRSM            & 3360 (train) + 840 (test) & 1920 (train) + 480 (test) \\
Extended DIRSM   & 9945 & 9625\\
                 & - 3360 (DIRSM, train)  & - 1920 (DIRSM, train)\\
                 & - 2000 (Two-class weather, cloudy)  & - 1206 (MediaEval'18, road not passable)\\
                 & - 2000 (Two-class weather, sunny)  & - 6499 (Collected Tweets/Instagram) \\
                 & - 2585 (Collected Tweets/Instagram) & \\
\bottomrule
\end{tabular*}
\label{tab:dataset}
\end{table*}

We evaluated the performance of our trained classifiers based on the DIRSM (Disaster Image Retrieval from Social Media) benchmark dataset offered by \textit{MediaEval'17 MMSat Task}~\citep{bischke2017multimedia}, where Flickr images were assigned with flood relevant or irrelevant annotations.

Since images from social media such as Twitter or Instagram may vary largely in quality, we introduced more annotated images from three additional data sources, namely our own annotated social media image collection, mainly from the flood in 2016 and 2017 in Europe~\citep{feng2018extraction}, 4000 randomly selected images from the two-class Weather Classification Dataset~\citep{lu2014two}, and the images annotated as containing scenarios where roads are not passable during the flood from \textit{MediaEval'18 MMSat Task}~\citep{bischke2018multimedia}. This dataset was named \textit{Extended DIRSM}. We built a relatively balanced dataset, which is beneficial for both training and evaluation. The distribution between training vs.\ test and positive vs.\ negative examples is summarized in Table \ref{tab:dataset}. 

\subsubsection{Experiment and Evaluation}

\begin{table}[width=.9\linewidth,cols=3,pos=h]
\caption{Evaluation of different approaches on \textit{MMSat Task} in \textit{MediaEval'17} and comparison with our approach}%\label{tbl1}
\begin{tabular*}{\tblwidth}{@{} LCC@{} }
\toprule
Methods & P@480 & AP@\big\{50,100, \\
 &  & 150,240,480\big\} \\
\midrule
%\cite{fu2017bmc}              & 15.55 & 19.69 \\
\cite{tkachenko2017wisc}      & 50.95 & 62.75 \\
\cite{zhao2017retrieving}     & 51.46 & 64.70 \\
\cite{lopez2017multi}         & 61.58 & 66.38 \\
\cite{hanif2017flood}         & 64.88 & 80.98 \\
\cite{nogueira2017data}       & 74.60 & 87.88 \\
\cite{dao2018context}         & 77.62 & 87.87 \\
\cite{avgerinakis2017visual}  & 78.82 & 92.27 \\
\cite{ahmad2017cnn}           & 84.94 & 95.11 \\
\cite{bischke2017detection}   & 86.64 & 95.71 \\
\cite{ahmad2017convolutional} & 86.81 & 95.73 \\
\midrule
our approach &&\\
\midrule
FC - InceptionV3      & 82.92 & 93.57 \\
FC - 3 models         & 87.08 & 97.25 \\
FC - 4 models         & 85.00 & 93.17 \\
Xgboost - InceptionV3 & 86.46 & 96.96 \\
Xgboost - 3 models    & \textbf{89.17} & \textbf{97.53} \\
Xgboost - 4 models    & 88.75 & 97.37 \\
\bottomrule
\end{tabular*}
\label{tab:compare_mediaeval17}
\end{table}

\begin{table*}[width=1.9\linewidth,pos=h]
\caption{Evaluation of model performance based on precision, recall and \Fone\ scores on positive class, Overall Accuracy (OA) and Area Under Curve (AUC)}%\label{tbl1}
\begin{tabular*}{\tblwidth}{@{} LLCCCCCCCCCCC@{} }
\toprule
Train set & Method & \multicolumn{5}{c}{DIRSM test set} & & \multicolumn{5}{c}{Ext. DIRSM test set} \\
& & Prec. & Rec. & \Fone & OA & AUC & & Prec. & Rec. & \Fone & OA & AUC\\
\midrule
DIRSM & 3 models - FC & 91.44 & 82.29 & 86.62 & 90.76 & 0.967 & & 95.53 & 70.50 & 81.13 & 83.60 & 0.950\\
DIRSM & 3 models - Xgboost & 89.31 & 88.75 & \textbf{89.03} & \textbf{92.05} & \textbf{0.972} & & 98.55 & 74.80 & \textbf{85.05} & \textbf{86.85} & \textbf{0.976}\\
\midrule
ext. DIRSM & 3 models - FC & 90.68 & 81.04 & 85.59 & 90.08 & 0.964 & & 94.22 & 86.40 & 90.14 & 90.55 & 0.972\\
ext. DIRSM & 3 models - Xgboost & 85.35 & 91.04 & \textbf{88.10} & \textbf{91.06} & \textbf{0.971} & & 92.51 & 92.60 & \textbf{92.55} & \textbf{92.55} & \textbf{0.982}\\
\bottomrule
\end{tabular*}
\label{tab:evaluation_testset}
\end{table*}

\begin{figure*}
\centering
\includegraphics[clip=true,trim=20pt 5pt 40pt 40pt, width=.49\linewidth]{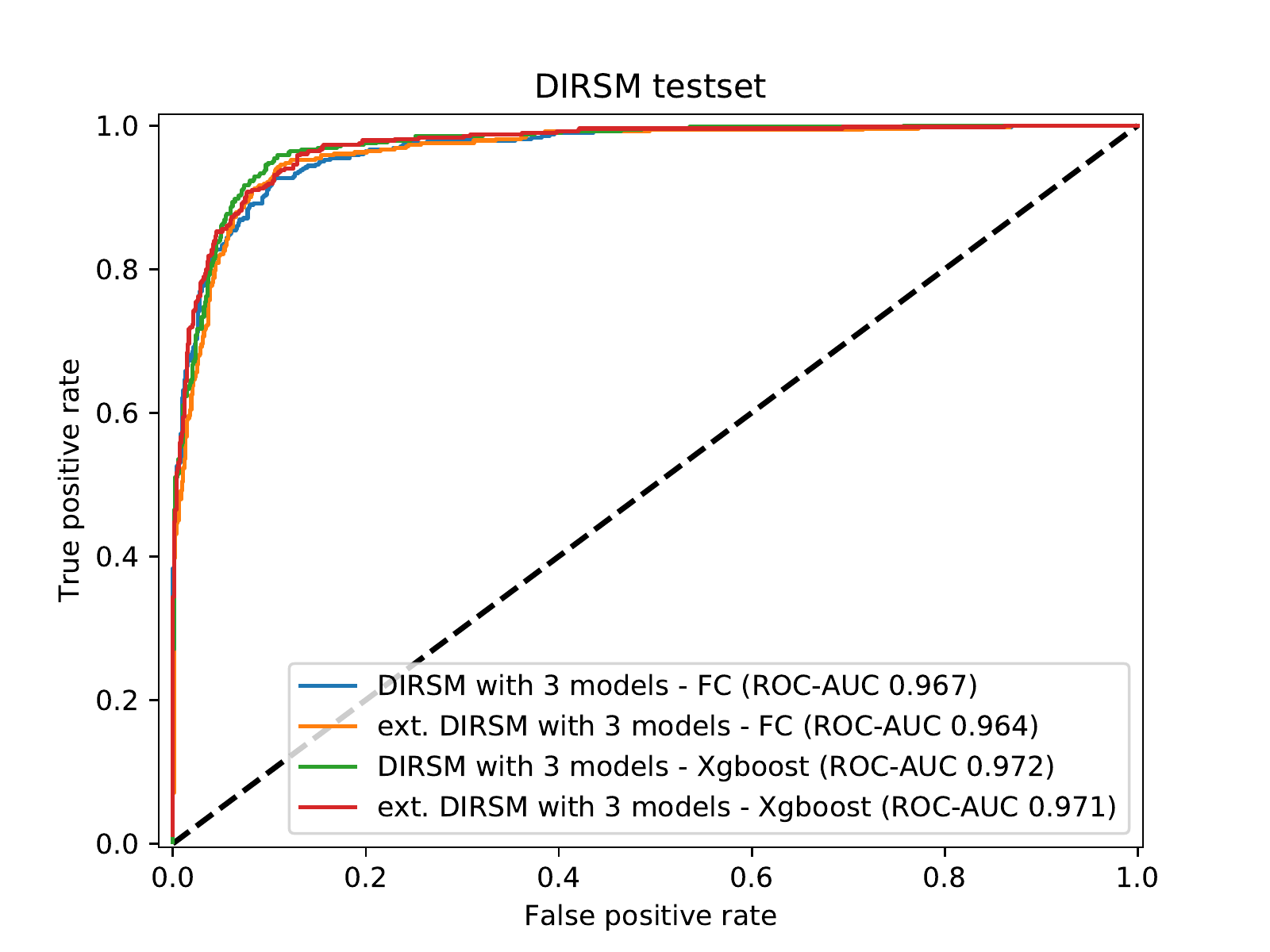}
\includegraphics[clip=true,trim=20pt 5pt 40pt 40pt, width=.49\linewidth]{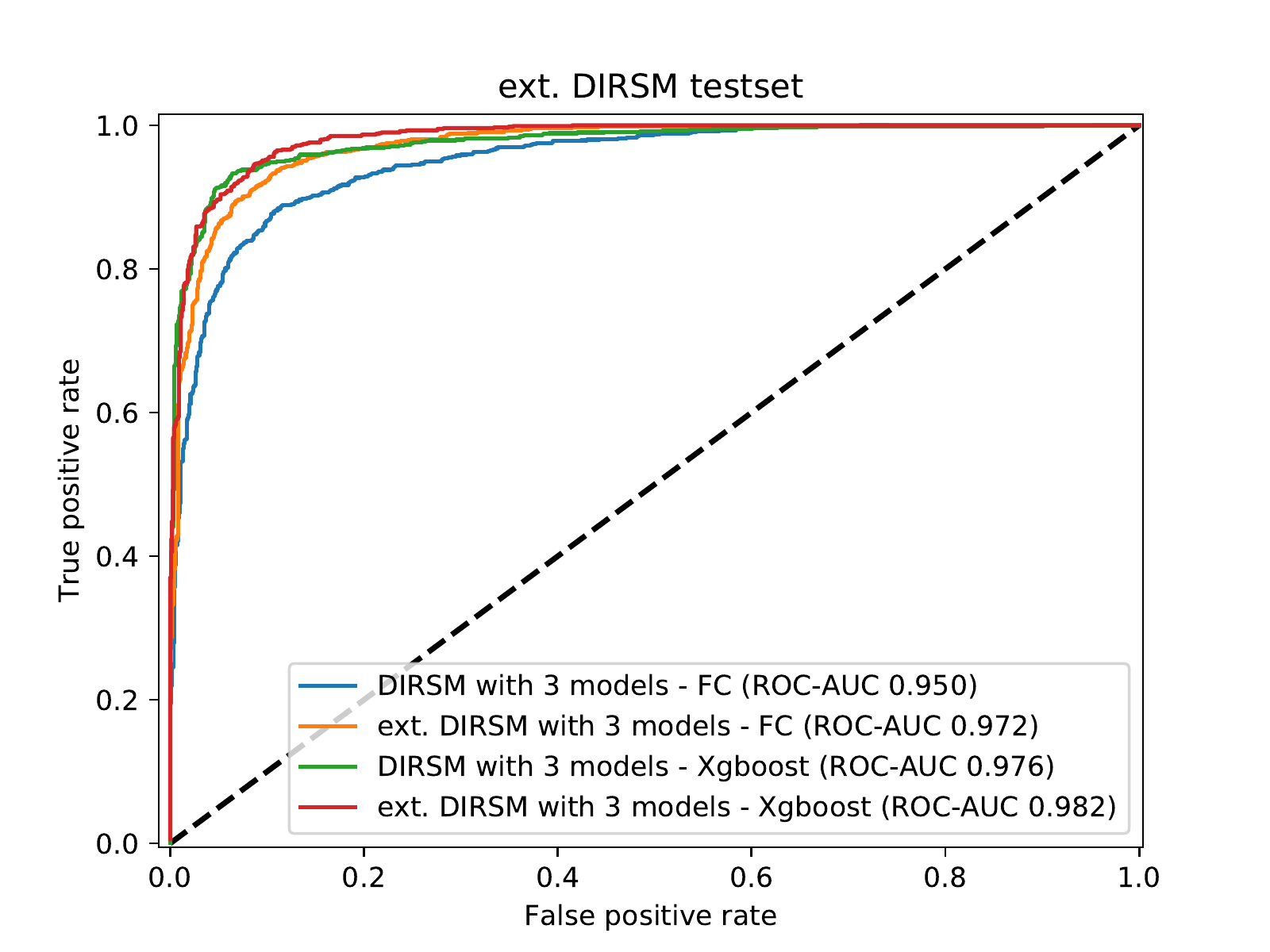}
\caption{Evaluation of models on DIRSM test set (left) and extended DIRSM test set (right)}
\label{fig:evaluation_testset}
\end{figure*}

Firstly, we trained the image classifiers with different combinations of pre-trained models, namely~\textit{InceptionV3} only, three models (containing~\textit{InceptionV3}, \textit{DenseNet201}, and \textit{InceptionResNetV2}), and four models (three models plus VGG16, pre-trained on Place365). Each combination was trained either with FC layers or Xgboost. 

The models were firstly evaluated based on the DIRSM dataset. In order to compare our result with existing work, same metrics as in those tasks were used. Since a ranking retrieval system is to be built, precision of the top-related documents is more relevant.
For this reason, cut-offs were applied on the ranked retrieval results and the precision was calculated. As the number of positive examples is 480, the metrics precision at cut-off 480 (P@480) and average precision at cut-offs 50, 100, 150, 240 and 480 (AP@\{50, 100, 150, 240, 480\}) were used for evaluation. We randomly selected 200 images from both the positive and negative training examples separately and used them as validation set. Early stopping with a patience of 6 epochs was applied when the validation loss did not constantly improve. The comparison with previous research using the same dataset is summarized in Table~\ref{tab:compare_mediaeval17}.
From the results, we conclude that the Xgboost classifier has generally outperformed the models using FC layers. The combination of three models achieves the best results and it also indicates that combining a VGG16 pre-trained on Place365 is not beneficial in our case. 

Secondly, in order to adapt to the larger variety of images from Twitter and Instagram, models were trained on the extended DIRSM dataset. We randomly picked 800 images from both classes as a validation dataset and 1000 images from both classes as the test set. Models were trained on the rest of the images. Since the combination of three models demonstrated the best performance, we used this strategy on both the DIRSM training set and extended DIRSM training set and then evaluated on both the DIRSM test set and extended DIRSM test set. 

Our purpose in this work differs slightly from a ranked retrieval system, as in the \textit{MMSat} task, since it is rather to reject off-topic posts efficiently. In this case, the false negative error plays a role. Thus, for the evaluation on the extended DIRSM dataset, we used the metrics such as precision, recall, \Fone-score, on the positive class, Overall Accuracy (OA) and Area Under Curve (AUC). The performance of the models is summarized in Table~\ref{tab:evaluation_testset}. The ROC curves of our trained models are compared in Figure~\ref{fig:evaluation_testset}.

From the evaluation on both test datasets, the Overall Accuracy and AUC of the Xgboost models are significantly higher. For the DIRSM test set, the benefits of introducing more annotated images are not obvious, however, both metrics are significantly improved on the extended DIRSM test set. This means that introducing more annotated images makes the classifier more adaptive to the images coming from Twitter or Instagram.

In summary, we trained a flood image classifier for social media images with state-of-the-art performance, which can filter out most of the off-topic images with an accuracy of 92.55\%.

%hier könnte das kapitel zur elimination von duplicates kommen - m.e. passt es sehr gut rein (ich lese es noch bis zum ende und überlege dann nochmals ob es passt ;-))

\subsection{Duplication detection}
In many cases, social media users may apply photo editing or add extra texts to others' images, therefore we cannot simply apply pixel-level comparison to detect such duplicates. Because of this, a deep feature based duplication detection was developed. Images were firstly processed to feature vectors with a pre-trained deep model. In this case, we used a light-weight model, ResNet18 \citep{he2016deep}, which generates 512 dimensional feature vectors from resized input images of $224\times224\times3$. The assumption is then, that similar images should also be close to each other in feature space, which can be revealed using clustering algorithms. In this work, we clustered the features using DBSCAN, a density-based clustering method.

\subsection{Water level estimation}
\label{sec:wlevel}
After filtering out off-topic images and removing duplicates, the third component is to estimate the water level, based on objects with known dimensions standing partly in the water. In this work, we selected people as our targets, because -- according to our observation -- they are the most common objects in social media image datasets. In the subsequent sections, we present the design of our proposed method and then evaluate it by comparison with two baseline methods.

\subsubsection{Learning a water depth classifier with handcrafted features}

According to our observation, the easiest way to determine the water level is to analyze an object of known size, which is partially covered by water. 
The relative proportions of human bodies are well known and thus a rough estimation of the parts covered by water can straightforwardly be determined -- as opposed, e.g. to buildings or vegetation. Thus, the task is to identify the body parts which are not covered by water.
In order to do so, we used three separate neural networks to provide the fundamental information. First, an object detection network detected people as bounding boxes. Using the second network, each person's body parts were identified. Finally, the third network is a segmentation network, which was used to provide the surrounding information around the persons.

The first neural network is Mask R-CNN~\citep{he2017mask}, which is one of the state-of-the-art frameworks for object detection. For each detected single object instance it outputs a class label and a bounding box. We used the \textit{Keras} implementation~\citep{matterport_maskrcnn_2017} of this network and applied the weights pre-trained on the MS COCO dataset \citep{lin2014microsoft}. The detection determines whether the image contains people which can be subsequently used for water level estimation.

\begin{figure}[pos=h]
	\centering
	\includegraphics[height=.5\columnwidth]{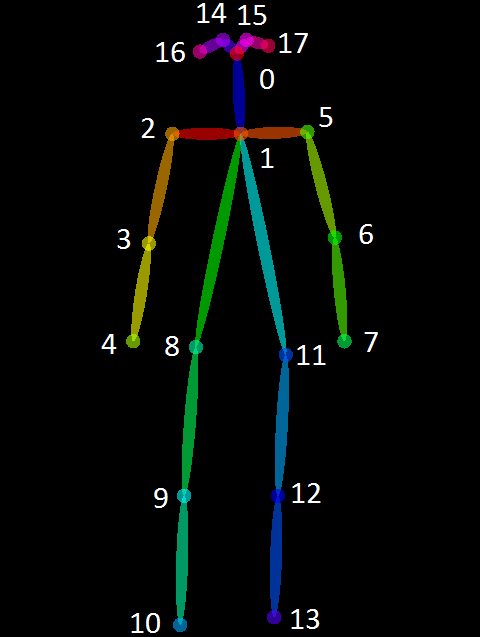}
	\caption{Output of OpenPose with 18 body keypoints \citep{openpose2018output}}
	\label{fig:openpose_output}
\end{figure}

Using the second neural network, people in the scene are detected and body keypoints are identified. In this work, we used OpenPose \citep{cao2018openpose} to detect multi-person keypoints. It is a multi-stage CNN, which cannot only provide the detected body keypoints but also their corresponding confidence scores.
The model detects 18 landmark points of the human body \citep{openpose2018output} as shown in Figure~\ref{fig:openpose_output}. Not all of the detected keypoints are relevant for water level estimation. Therefore, we selected only the keypoints 0, 1, and 8-13 to represent the human body, and neglected the keypoints of arms and eyes.
%Please note that always all keypoints are fitted to an identified human shape - also the ones which are occluded or not visible. This is an important aspect of our approach.

The third neural network aims at the identification of surrounding pixels of a person by image semantic segmentation. Especially, we focused on two classes, namely ground and water. In this work, we used Deeplabv3+ \citep{deeplabv3plus2018}, which is one of the state-of-the-art architectures for semantic segmentation.
Specifically, we used a Deeplabv3+ network pre-trained on the ADE20K dataset \citep{tensorflow2019deepLab} for semantic image segmentation, which achieves a 82.52\% pixel-wise accuracy on the ADE20K validation set.
The ADE20K dataset \citep{zhou2017scene} was annotated with more than 250 classes, which include the two classes we focused on.
% Model name: xception65_ade20k_train

\begin{figure*}
		\centering
		\includegraphics[clip=true,trim=30pt 130pt 200pt 40pt, width=1.7\columnwidth]{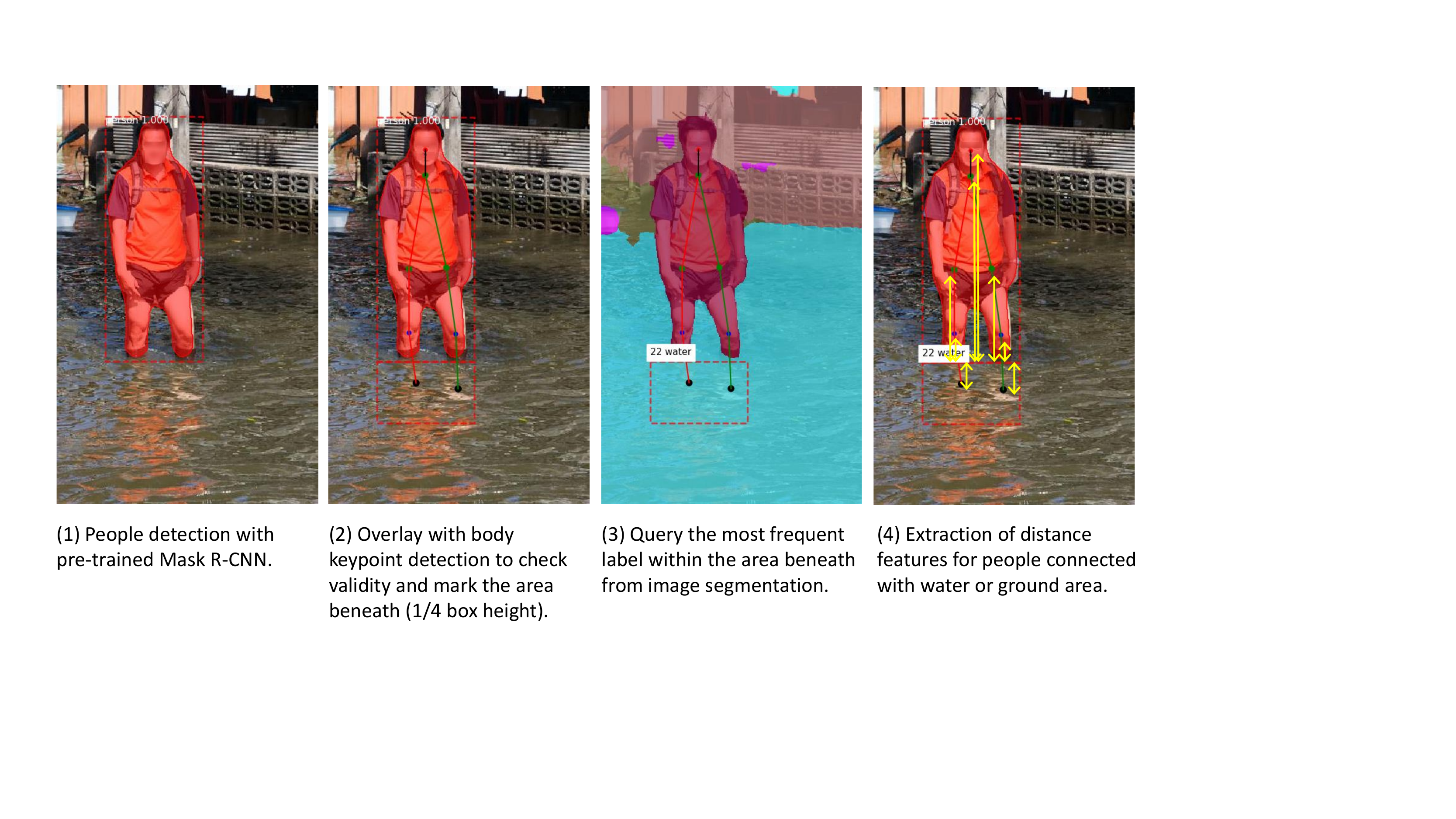}
	\caption{Steps for extracting handcrafted distance features (example image under \href{https://creativecommons.org/licenses/by-nc-sa/2.0/}{CC BY-NC-SA 2.0})}
	\label{fig:handcrafted_features}
\end{figure*}

With outputs from the above-mentioned models, it is possible to determine the water height relative to the human body. The difference between the bounding box of the human shape and the keypoints indicates the hidden body parts. 
In order to determine the water level, a classifier was built, which is based on a feature vector, created by the sequence of steps shown in Figure \ref{fig:handcrafted_features}. 

First, the pre-trained Mask R-CNN was used to detect people, resulting in bounding boxes. Then, body keypoints were overlaid, and only the bounding boxes with corresponding body keypoint detections were preserved. In this way, we only retained the people, which could be detected by both, the object detection model and the body keypoint model, for further analysis.

In the third step, we hypothesized the waterline to be at the bottom line of the bounding box of a person, which was detected by the Mask R-CNN. 
%To verify this hypothesis, 
An area beneath the bounding box with a box height of 1/4 of the given bounding box is marked. In this box, the most frequent class label from the segmentation results is queried (e.g. water, ground, but also classes such as cars, boats). We kept only the people connected to an area of ground or water. Preservation of ground is necessary, as the segmentation algorithm detects water segments only in case of a severe flood. For most of the other cases (e.g. ankle level flood), flooded areas are mostly predicted as being ground. Thus, we considered both classes -- ground and water, as our focus classes.

Lastly, a distance feature vector $\mathbf{D_{box\_bottom}}$ was calculated from the water line (box bottom) to all used keypoints (8 values, see Figure \ref{fig:handcrafted_features}). These distances were further normalized by the box height to eliminate the
% perspective
% [CB: I think just the scale, not the
%  perspective mapping]
influence of the unknown scale,
\begin{eqnarray}\label{3}
\mathbf{D_{box\_bottom}} = \frac{y_{bottom} - \mathbf{Y_{keypoints}}}{y_{bottom} - y_{top}}
\end{eqnarray}

\noindent where $\mathbf{Y_{keypoints}}$ is a vector of y-coordinates of all the used keypoints in the image coordinate system, $y_{top}$ and $y_{bottom}$ are y-coordinates of the top and bottom line of the bounding box. 
These built the Feature Group~1 (FG~1). 

We further considered two additional groups of features: Feature Group~2 (FG~2) contains the confidence scores of the OpenPose keypoint detection (8 values), which indicate how well each keypoint can be detected. Lastly, Feature Group~3 (FG~3) is a binary value, which indicates whether the person is connected to a water area or a ground area.

Thus, in total, a feature vector of 17 values was used to represent one person, consisting of two feature groups of 8 values each, and one additional binary feature. Then, a classic machine learning method, such as SVM (Support Vector Machine) or random forest, could be applied to determine the water height relative to the body frame, in terms of the water level classes \textit{ankle}, \textit{knee}, \textit{hip}, \textit{chest} and in addition, \textit{no evidence}. In this work, we opted to use the more state-of-the-art classifier Xgboost \citep{chen2016xgboost} for training the water level estimation model. 

\subsubsection{Dataset}
\label{sec:water_level_dataset}

\begin{figure}[pos=t]
	\centering
	\includegraphics[height=.55\columnwidth]{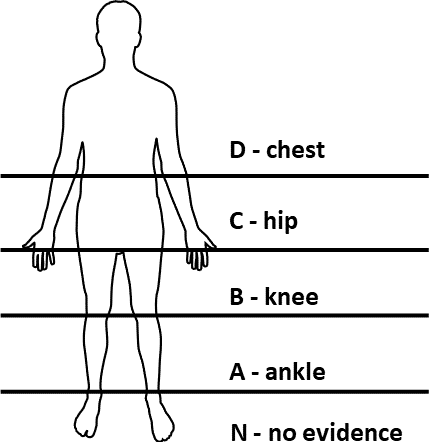}
	\caption{Annotation rules for water level estimation of single person}
	\label{fig:annotation}
\end{figure}

To the best of our knowledge, there is no public dataset or benchmark available for this task. The only comparable dataset which appeared in previous research is used for the work of~\cite{chaudhary2019flood}, where 7000 images were annotated pixel-wise into 11 water level classes. The images were collected from various Internet sources, such as news articles, search engines, and social media. However, it is not yet publicly available. 
Therefore, we collected similar data sources with images from flooding or heavy rainfall scenarios which contain at least one person. As the final goal of this work is to provide a water level estimation for each geotagged social media image, we are interested only in estimating \textit{one} water level for each image. Thus, there is no need to provide pixel-wise labels for every image. 
For this reason, instead of an annotation of all image pixels, we annotated the whole image with one single label, which is much less time-consuming. Regarding the case when multiple people stand in the water (in different heights), these images were annotated with the label of the majority. We annotated the images into five classes with the rules shown in Figure \ref{fig:annotation}. \textit{N} stands for all persons who have no evidence for water level estimation, e.g.\ standing on wet ground, standing on the river bank, or sitting in a boat. From \textit{A} to \textit{D}, the label is associated with the water level at ankle, knee, hip and chest. The images were annotated according to this rule by one annotator.  

We collected 1375 images containing persons in flood or heavy rainfall situations. They were annotated into the above mentioned 5 classes. Additional 325 images from the MS COCO dataset were introduced as class \textit{N}, which contains mostly the situation of people standing on the ground with no significant water level. From each class, 50 images were randomly selected as the test set and kept unseen during training. In total, 1700 images were used as our dataset; the composition of the dataset is summarized in Table \ref{tab:data_composition}.

\begin{table}[width=.9\linewidth,cols=3,pos=h]
	\caption{Composition of train set and test set for water level estimation}
	\begin{tabular*}{\tblwidth}{@{} LCC@{} }
		\toprule
		Class Name & Train Set & Test Set \\
		\midrule
		N - No evidence & 450 & 50 \\
		A - Ankle & 250 & 50 \\
		B - Knee  & 250 & 50 \\
		C - Hip  & 250 & 50 \\
		D - Chest & 200 & 50 \\
		\bottomrule
	\end{tabular*}
	\label{tab:data_composition}
\end{table}

%The amount is about 1/4 of the image containing flood scenarios. In total 1676 images were used as a development set, where 20\% of them were used as a validation set. An independent test set, which has no duplication to the development set was used for evaluation.

\subsubsection{Pseudo Labelling}
\label{sec:pseudo_labeling}
The annotations of the flood levels are per image, while our water depth estimation is per instance (i.e., each person in the image). This creates a potential problem, since simply assigning the image level annotations to each instance may mislead the training process. As an example, an image may show several people standing in different water levels, while some others are sitting in boats.
Thus, we can regard this as a Multiclass Multiple Instance Learning (MIL) problem. All images are regarded as bags of instances. Only the annotations of the bags are given. The model, however, needs to predict each instance in the image. One of the possible strategies is pseudo labelling. We assigned the bag annotation to each instance in the image and trained a model. The instance level annotations were updated based on the confidence score of the softmax outputs. 
If the confidence score is above a relatively high value (0.85 in this work), it means the model is very sure about its prediction. Thus, we replaced this instance label with its predicted label and trained this model again. This step was repeated until no further updates happened for the instance level annotations. 

Another issue is the reasoning of the final prediction for the whole image. We firstly neglected the persons classified as \textit{N} by our classifier. 
In the case when all of the persons were neglected, the final prediction of the image is \textit{N}.
For the remaining persons, we used the majority to make the final prediction for the image.
%we voted for a final prediction of the image
If the majority votes are equal, we took the prediction with a higher confidence score based on the softmax output.
% The result of all instance predictions is analyzed: it is either the unanimous result, or the prediction with the highest confidence score based on the softmax output is taken.

\subsubsection{Baseline 1: Multiclass image classification with global deep features of the whole image}
\label{sec:baseline1}
For a comparison with our proposed method, we applied a simple multiclass classification using global deep features as baseline. The same late fusion architecture as described in Section~\ref{sec:retrieval_method} was applied, where features generated by pre-trained \textit{DenseNet201}, \textit{InceptionV3} and \textit{InceptionResNetV2} on ImageNet were concatenated and then classified with Xgboost. Instead of a binary classification, softmax outputs were generated for all five water level classes. From this, the performance of this model indicates, whether the global deep features are beneficial for water level estimation.

\subsubsection{Baseline 2: Mask R-CNN with extra branch for water level classification}
\label{sec:baseline2}

\begin{figure}[pos=t]
	\centering
		\includegraphics[clip=true,trim=57pt 60pt 235pt 130pt, width=1.\columnwidth]{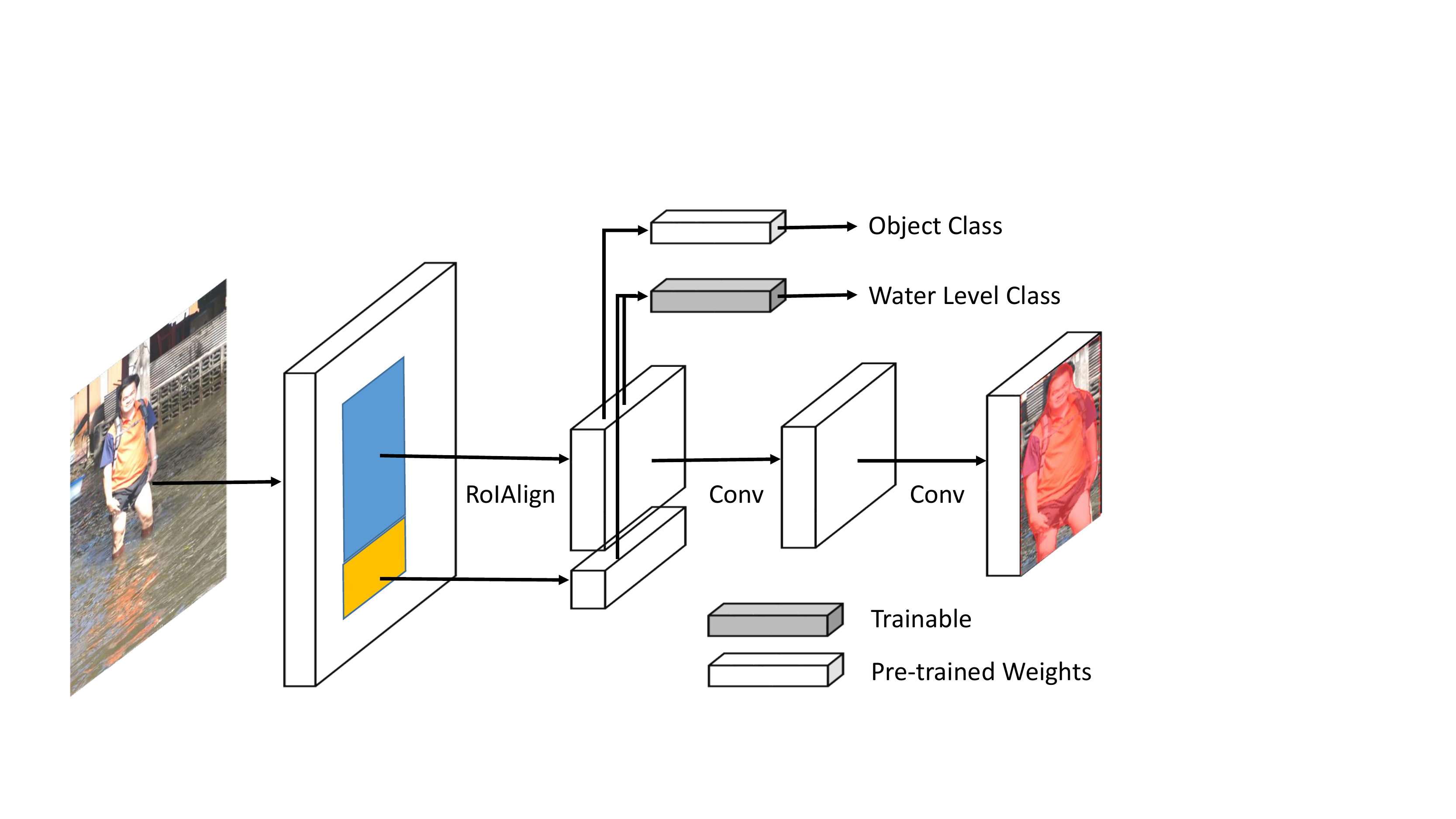}
	\caption{Network architecture of baseline 2: Mask R-CNN with water level classification branch using local deep features (example image under \href{https://creativecommons.org/licenses/by-nc-sa/2.0/}{CC BY-NC-SA 2.0})}
	\label{fig:flood_mask_rcnn}
\end{figure}

In order to classify the water level based on the local deep features around each person, the implementation of Mask R-CNN with \textit{Keras} provided by \citep{matterport_maskrcnn_2017} was extended with an extra classification branch for water level classification as the second baseline.
We used the default parameter settings for Mask R-CNN. A backbone network, ResNet101, was used for extracting deep features at different spatial scale, which is also known as FPN (Feature Pyramid Network). The RPN (Region Proposal Network), mask branch, classification branch, box branch were trained based on the feature maps generated from FPN separately. We added an extra branch which is the same as the classification branch for water level estimation. It classified with a cross-entropy loss based on the output of FPN. For the original parts, such as FPN, RPN, box branch and classification branch, we initialized with the weights pre-trained on the MS COCO dataset. We froze the object detection parts of the network and trained the custom water level classification branch on our dataset.
Furthermore, as we noticed that considering the area below the detected persons might contribute to the water level classification, an adapted version of this network architecture fed both the FPN outputs from the object area and the area of 1/4 of the box height beneath the object to the water level classification branch. The network architecture is shown in Figure \ref{fig:flood_mask_rcnn}.

The idea of this baseline is similar to \cite{chaudhary2019flood}. The main difference is that they trained the model from scratch based on pixel-level annotations of their flood level dataset and part of the MS COCO dataset, whereas this baseline inherits the object detection function directly from pre-trained weights. Our flood level dataset was only used for tuning the FPN and training the water level classification branch. Therefore, we only need to provide labels for each person instance and the model can be trained with the pseudo labelling strategy with one single 
%global or: single, unique? instead of global? 
label for the whole image as described Section \ref{sec:pseudo_labeling}. 
Thus, our annotation cost for this baseline is much less. Additionally, they used not only people, but also many other object classes, such as houses and cars, whereas our dataset is annotated only based on the water level measure of persons. Even though this baseline is not the same as the work from \cite{chaudhary2019flood}, it generally represents the ability of water level classification, which makes use of the local deep features around detected persons.

\subsubsection{Experiment and evaluation of water level estimation}
\label{sec:experiment_and_evaluation}

\begin{table}[width=.9\linewidth,cols=3,pos=h]
\caption{Parameters for all methods}%\label{tbl1}
\begin{tabular*}{\tblwidth}{@{} LL@{} }
\toprule
Method  & Parameters \\
\midrule
Ours & \textbf{Xgboost} \{max-depth:2, eta:0.3, \\
 & objective:multi-softmax, silent:1, \\
 & num-class:5, num-round:300, \\
 & early-stopping-rounds:20\} \\
Baseline 1 & \textbf{Xgboost} \{max-depth:2, eta:0.3, \\
 & objective:multi-softmax, silent:1, \\
 & num-class:5, num-round:300, \\
 & early-stopping-rounds:20\} \\
Baseline 2 & \textbf{Mask r-cnn} \{batch-size:1,\\
 & max num epochs:80, steps per epoch:300,\\
 & learning rate:0.00005, \\
 & early-stopping patience: 10\} \\
\bottomrule
\end{tabular*}
\label{tab:parameters}
\end{table}

Our proposed model and the two baselines, were trained on the same data set, where 20\% of the data were used for validation and the rest for training. 
During the experiments, it was observed that many of the wrong predictions were due to the very small size of people at far distances. Therefore, it is required that the number of pixels of the detected people segments must be greater than 0.1\% of the total pixel number of the whole image. Some of the important parameters used for training the models are listed in Table \ref{tab:parameters}.

\begin{figure}[pos=h]
	\centering
	\includegraphics[clip=true,trim=38pt 110pt 32pt 32pt, width=.9\columnwidth]{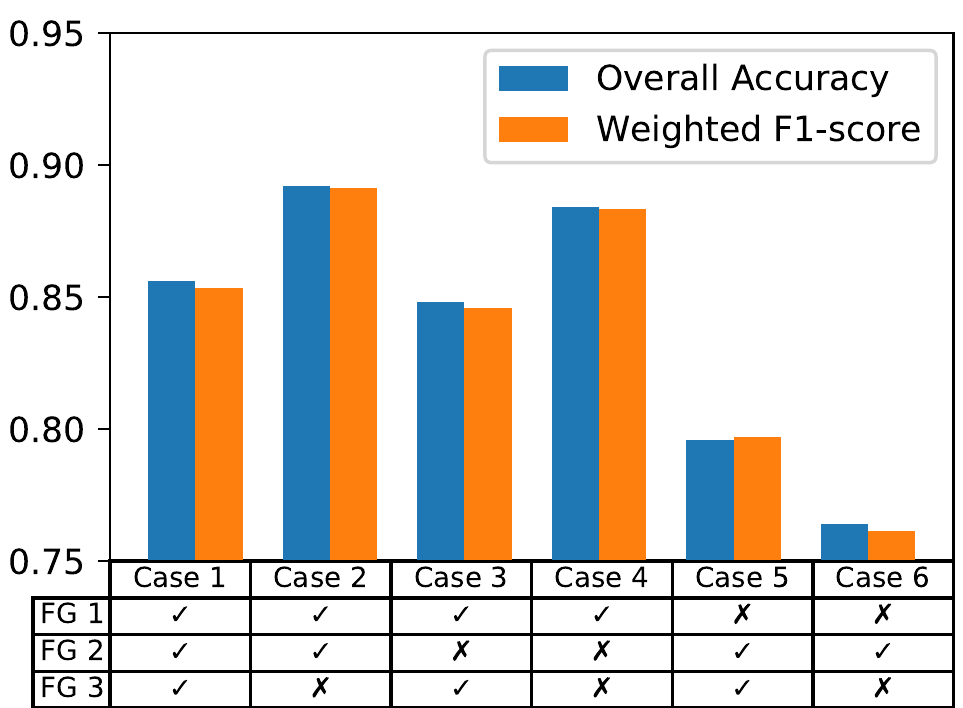}
	\caption{Evaluation of different combinations of feature groups performed on test set}
	\label{fig:feature_evaluation}
\end{figure}

\begin{figure*}
	\centering
	\includegraphics[clip=true,trim=45pt 240pt 45pt 5pt, width=2.05\columnwidth]{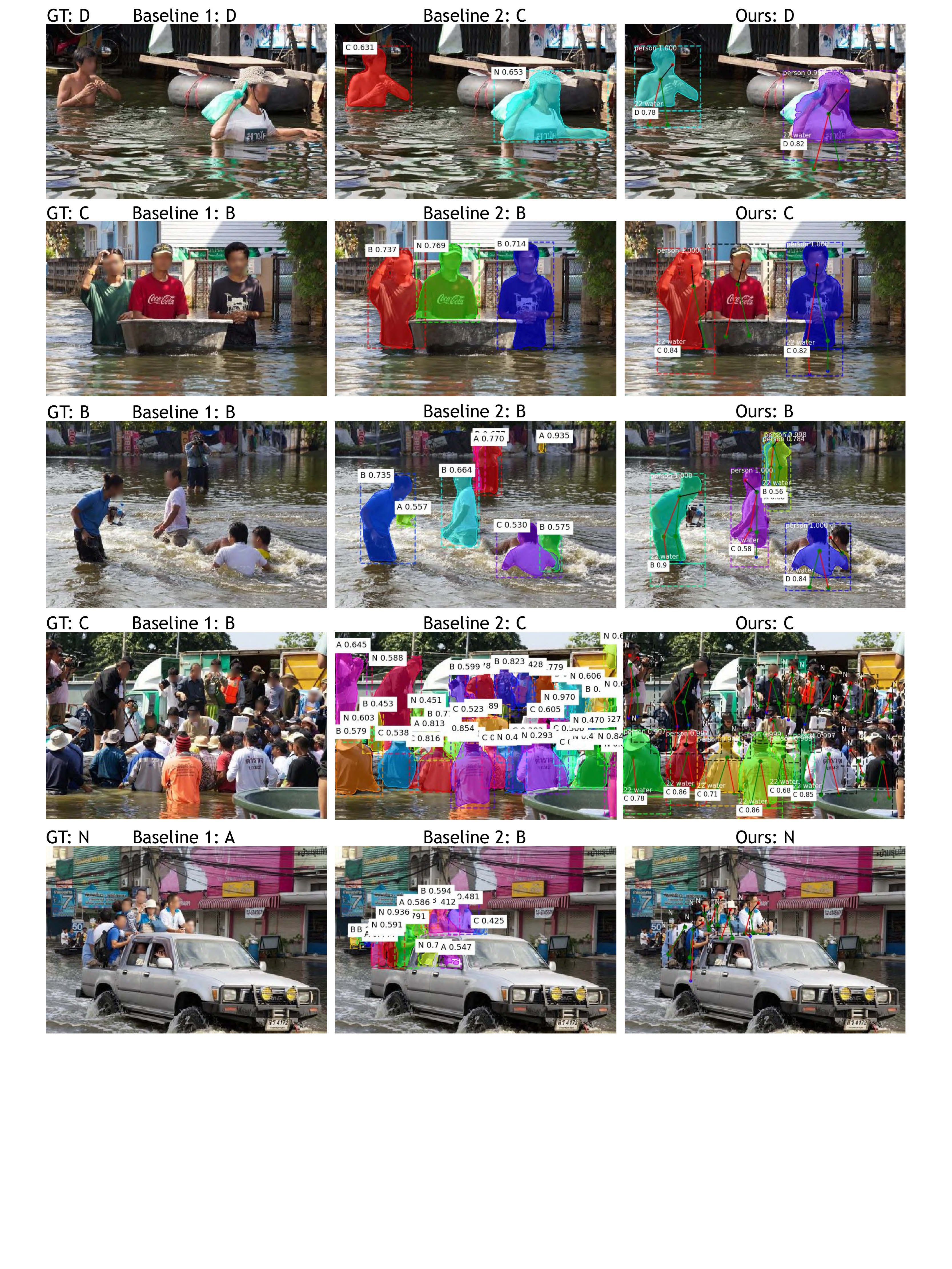}
	\caption{Qualitative evaluation of our proposed approach compared with the baselines (example images under \href{https://creativecommons.org/licenses/by-nc-sa/2.0/}{CC BY-NC-SA 2.0})}
	\label{fig:qualitative}
\end{figure*}

\begin{figure*}
	\centering
	\includegraphics[clip=true,trim=25pt 1070pt 60pt 30pt, width=2.05\columnwidth]{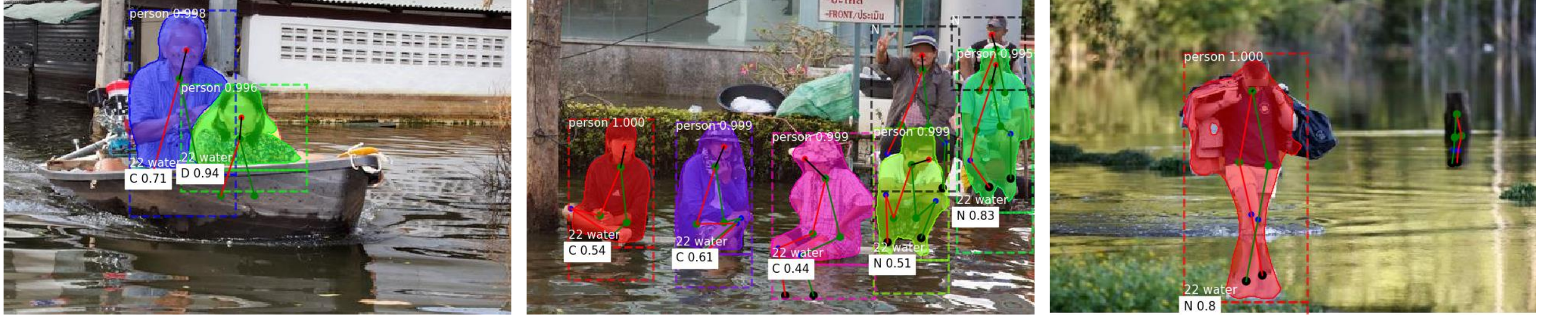}
	\caption{Example failure cases of our approach, caused by segmentation failure - left, sitting people - middle, and water reflection - right (example images under \href{https://creativecommons.org/licenses/by-nc-sa/2.0/}{CC BY-NC-SA 2.0})}
	\label{fig:failed}
\end{figure*}

\begin{figure*}
\centering
\includegraphics[clip=true, trim=0pt 0pt 0pt 13pt, width=.32\linewidth]{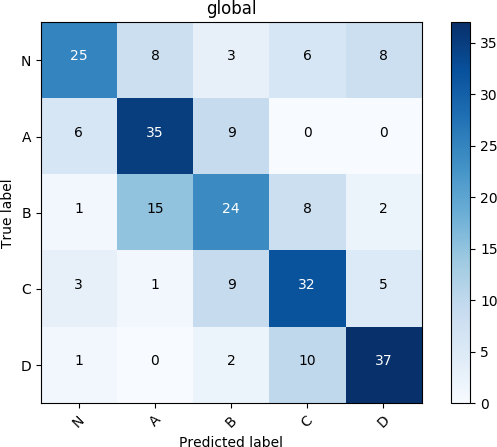}
\includegraphics[clip=true, trim=0pt 0pt 0pt 13pt, width=.32\linewidth]{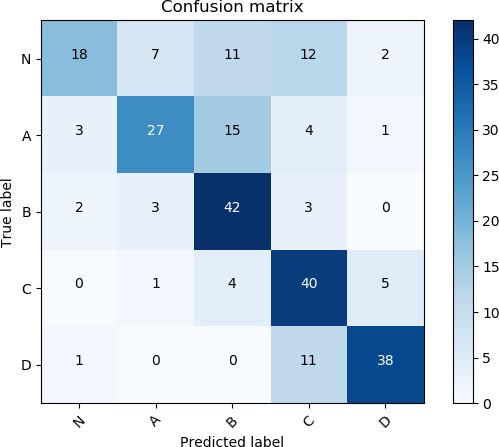}
\includegraphics[clip=true, trim=0pt 0pt 0pt 13pt, width=.32\linewidth]{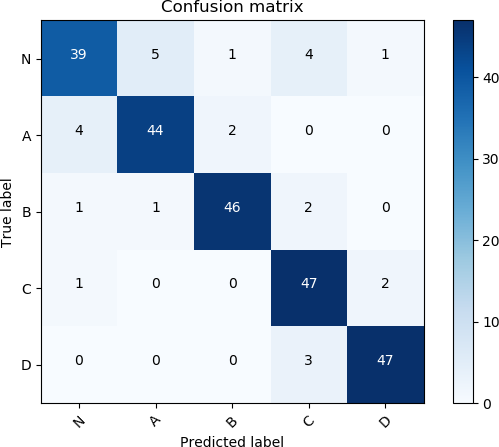}
\caption{Comparison of confusion matrices on the test set using baseline 1 (left), baseline 2 using 1/4 area beneath (middle) and our proposed method (right)}
\label{fig:evaluation_baselines}
\end{figure*}

\begin{table*}[width=1.9\linewidth,pos=h]
\caption{Quantitative comparison of models for water level estimation}\label{eval_water_level}
\begin{tabular*}{\tblwidth}{@{} LCC@{} }
\toprule
Method & Overall Accuracy & Weighted \Fone-score \\
\midrule
Baseline 1 - global deep features& 61.20\% & 60.95\%  \\
Baseline 2 - adapted Mask R-CNN with no area beneath & 57.60\% & 52.94\%  \\ 
Baseline 2 - adapted Mask R-CNN with 1/4 area beneath & 66.00\% & 64.94\%  \\
Ours - handcrafted distance features & 89.20\% & 89.14\%  \\
Our model fused with baseline 1 & \textbf{90.00\%} & \textbf{90.01\%}  \\
\bottomrule
\end{tabular*}
\label{tab:evaluation_baselines}
\end{table*}

Firstly, we analyzed all the combinations of the three feature groups for our proposed method as shown in Figure \ref{fig:feature_evaluation}. The model was trained with different feature groups separately. The overall accuracy and weighted \Fone-score on the test set was used as the performance measure. 
It is identified that FG~1 (distances to bounding box bottom) plays an important role in the classification, and a significant performance drop can be observed when FG~1 is excluded (see cases 5 and 6). For all cases using FG~1, a performance of over 85\% has been achieved. For most of the cases, including FG~3 (binary label for whether the connecting area is water or ground) is less beneficial. Combining FG~2 (OpenPose confidence scores) can slightly improve performance. Lastly, it is observed that the combination of FG~1 and FG~2 achieves the best results. 
Therefore, we used this strategy to train our best model and compared it with the two baseline methods described in Section \ref{sec:baseline1} and \ref{sec:baseline2}.

Secondly, some qualitative evaluations are shown in Figure \ref{fig:qualitative}, where five example images are presented with different water levels. The example images were collected from the Flickr album ``Flood - Thailand'' \citep{ebv2011flood}, published under CC BY-NC-SA 2.0 license. These images were kept unseen during the training of our models.
The ground truth (GT) and predictions for each image from our model together with the baselines are given. From the results, it can be observed that our model can ignore the majority of the persons showing no evidence to water level. Based on the bounding box, the features can present the proportion of visible and non-visible body parts. 
In baseline 2, the water level estimations contain many wrong predictions, especially for the people showing no evidence to water level. Baseline 1 predicts a knee level flooding more frequently, and also cannot distinguish images showing no evidence of water level properly.

Additionally, some failed cases of our approach are presented in Figure \ref{fig:failed}. In general, they are three common situations. On the left image, the segmentation network cannot provide a reliable prediction as the boat pixels are mostly predicted as water in this image. Therefore, these people have been classified as standing in the water to the hip (C) or chest (D) level. Sitting people in the water can also hardly provide reliable evidence for water level estimation. As the example shown in the middle, the three sitting people on the left hand side cannot be rejected properly. It leads to a wrong prediction of this image.
The third failure case on the right is caused by water reflection, where both the object detection and body keypoints estimation failed.

Thirdly, the quantitative analysis is presented, as shown in Table \ref{tab:evaluation_baselines} and Figure \ref{fig:evaluation_baselines}, where the confusion matrix, overall accuracy, and weighted \Fone-score are presented for the best model from different experiment settings. Analyzing the results reveals that our method achieves the best performance, compared to the two baselines. According to the confusion matrices, more examples are located at the diagonal of the matrix. It achieves over 89\% accuracy and weighted \Fone-score on our test set of 250 images. 
Baseline 1 has in general difficulties distinguishing neighbouring water levels.
Baseline 2 can be improved by introducing the features from the area beneath the detected box, however, it is still not as good as our proposed method. 
There are many images which were assigned with water level labels, even though there is no evidence for flooding.
In summary, our proposed method is a suitable solution for water level estimation and can be used for the flood severity mapping.

Furthermore, we considered fusing the softmax outputs from our model using hand-crafted features and baseline 1 using global deep features. Our model makes the final decision based on voting, thus the person predicted as the voted result with the highest confidence score according to softmax outputs is selected for fusion. Both the softmax outputs from the two models are linearly combined with weights.
Empirically, with weights 0.5 for our model and 0.5 for baseline 1, we were able to achieve a slightly higher model accuracy and weighted \Fone-score of 90\% on our test set.

%%%%%%%%%%%%%%%%%%%%%%%%%%%%%%%%%%%%%%%%%%%%%%%%%%%%%%%%
%                  % Model 2  Correct % Model 2  Wrong %
% Model 1  Correct %        132       %       75       %
% Model 1  Wrong   %        21        %       22       %
%%%%%%%%%%%%%%%%%%%%%%%%%%%%%%%%%%%%%%%%%%%%%%%%%%%%%%%%

% \FloatBarrier 
\section{Flood severity mapping for Hurricane Harvey in 2017}
\label{sec:case_study}
In order to show the benefits of our approach for flood severity mapping, we applied the proposed process on a severe flood event caused by Hurricane Harvey in 2017. Together with Hurricane Katrina in 2005, Hurricane Harvey was regarded as the costliest hurricanes by National Hurricane Center, NOAA \citep{noaa2018costliest}. It inflicted a damage of \$125 billion in Southeast Texas, especially the Houston metropolitan area. The strong precipitation led to a severe flood in the Houston area from 25\textsuperscript{th} of August to the 1\textsuperscript{st} of September 2017. Many studies have been conducted by researchers and national agencies in the last few years, which can provide much additional information for comparison and discussion.

The framework developed in our previous research \citep{feng2018extraction} has been constantly used for collecting Tweets via Twitter API in 2017. In order to eliminate the access limit of this API, we collected Tweets for the east, middle and west of the United States separately. Spatially, our data collection covered the whole disastrous area, and temporally, it covered all 8 days with significant flooding events. From 25\textsuperscript{th} of August to the 1\textsuperscript{st} of September 2017, we retrieved in total 150,227 Tweets with either geo-coordinates or location information in the Houston area. 28,833 of them contained URLs for photos; the photos were, however, not downloaded at that time. After deleting duplicate messages based on identical texts, 20,399 unique Tweets were retrieved. 
%Since many users may delete or modify the privacy setting of their posts, and also many users might upload multiple images at once, 
Two years later, on 13\textsuperscript{th} of June 2019, we were able to download 20,824 valid images for further image analysis. In the following, the application of the proposed process is presented, followed by the visualization of three mapping possibilities presenting the extracted information.

\begin{figure}
	\centering
	\includegraphics[clip=true,trim=30pt 10pt 502pt 25pt, width=.9\linewidth]{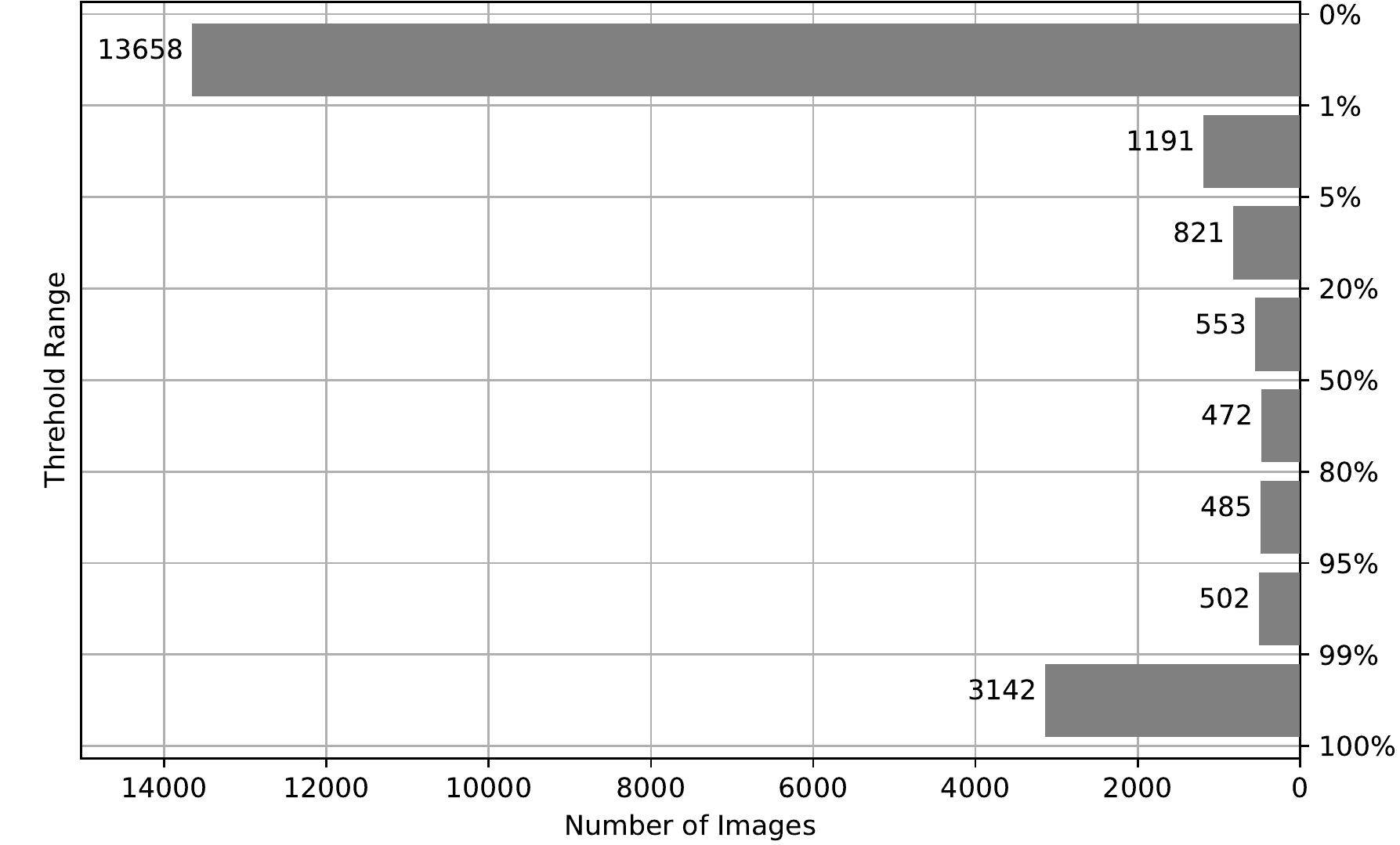} % threshold in percentage
	\caption{Distribution of the model predicted flood relevance scores for the images collected during Hurricane Harvey}
	\label{fig:filtering_results}
\end{figure}

\subsection{Processing of social media images}

Social media users may share images copied or duplicated from others. As such images often demonstrate severe flood situations and seemingly appear at multiple locations in a city, they can significantly mislead the mapping results. Therefore, the detection of duplicated images is an essential step before flood mapping. Thus, the processing of social media images has the following three steps in this application, (1) retrieve the flood relevant images, (2) remove duplicates of the images predicted as relevant, and (3) estimate the flood severity from the image collections.

\subsubsection{Social media filtering}

The binary classifier as trained in Section \ref{sec:retrieval} was applied on all downloaded images to retrieve the ones relevant to flood events. Since our model can provide an output with confidence score, we categorized the images into eight predefined groups with the thresholds 99\%, 95\%, 80\%, 50\%, 20\%, 5\% and 1\%, as visualized in Figure \ref{fig:filtering_results}. As shown in the bar diagram, 13,658 (65.6\%) of the collected images are surely irrelevant to the flood event, while 3,142 (15.1\%) are relevant; uncertainty exists for the remaining 19.3\% of the images.

\subsubsection{Duplication detection with deep features}

\begin{figure}
	\centering
	\includegraphics[clip=true,trim=0pt 10pt 0pt 40pt,width=.95\columnwidth]{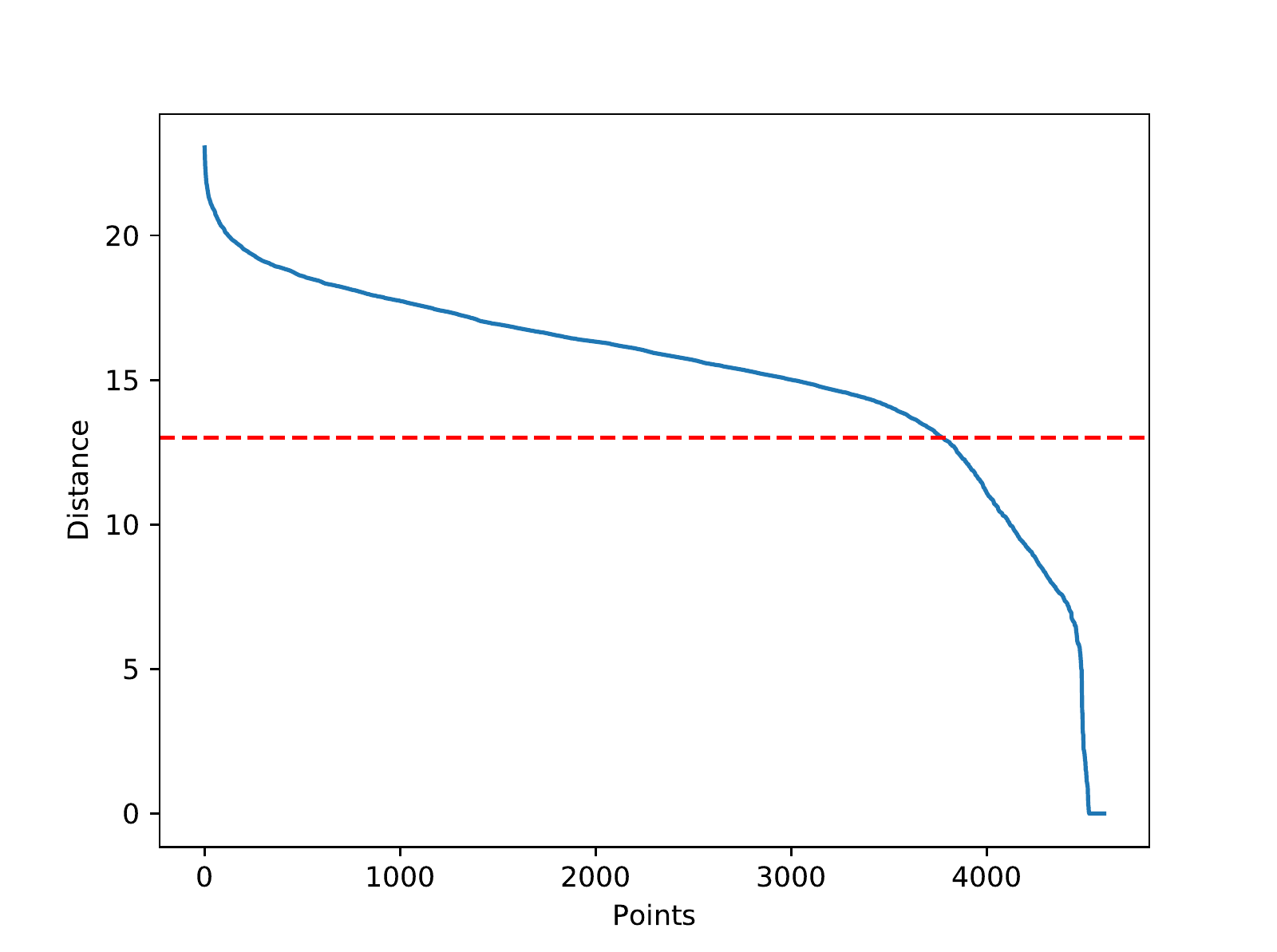}
	\caption{Sorted 2-distance plot for image deep features}
	\label{fig:knn}
\end{figure}
\begin{figure}
	\centering
	\includegraphics[clip=true,trim=100pt 70pt 310pt 20pt,width=.75\columnwidth]{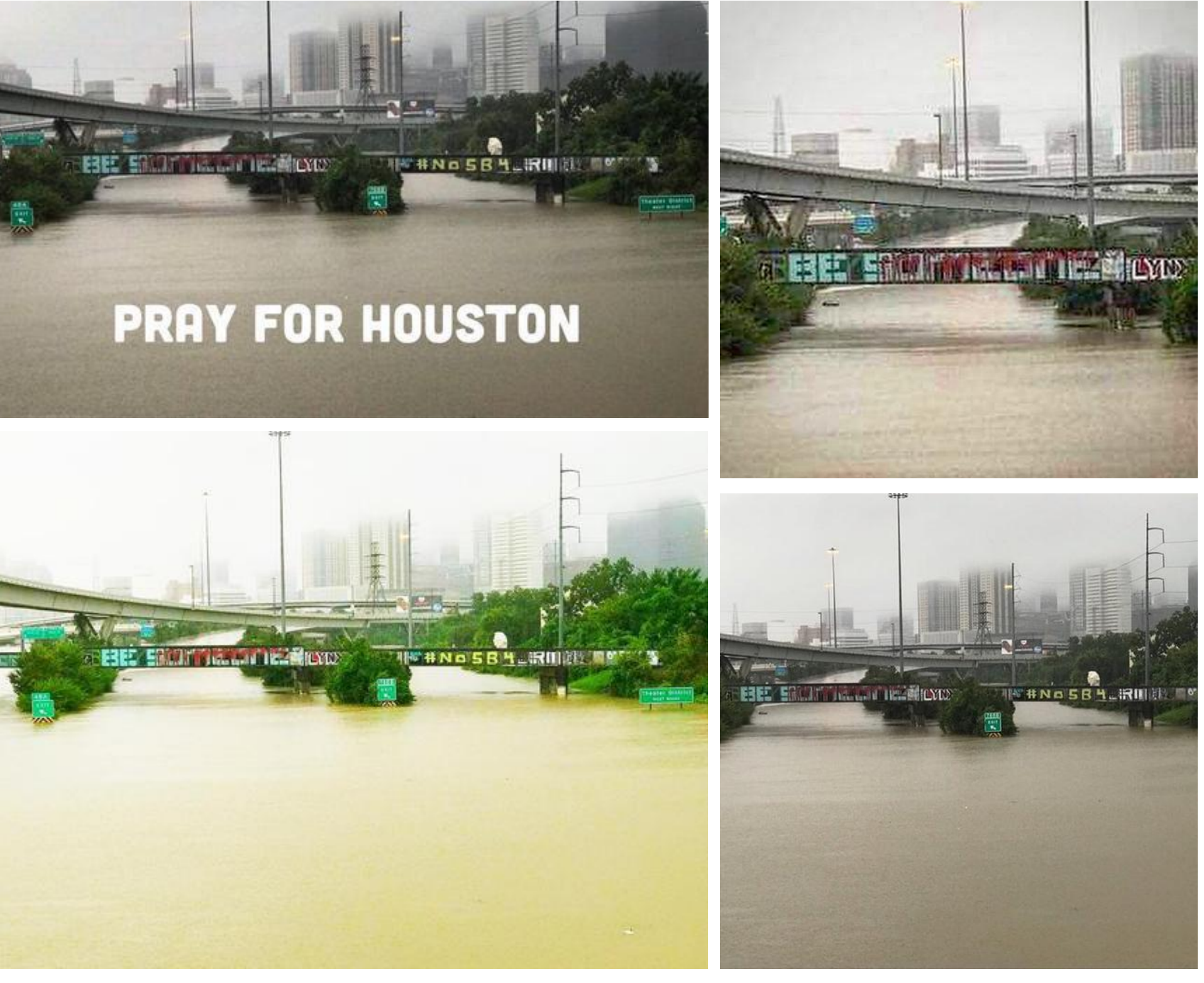}
	\caption{Examples of the duplicated images from the greatest cluster of DBSCAN result}
	\label{fig:duplicate}
\end{figure}

The duplication detection was applied to the flood relevant images, where we applied a 50\% threshold to the confidence score of the outputs. A total of 4,601 images were used for duplication detection.
Feature vectors were first generated with a pre-trained ResNet18 and then clustered using DBSCAN.
%
%In many cases, social media users may apply photo editing or add extra texts to others' images, therefore we cannot simply apply pixel-level comparison to detect such duplication. Because of this, a deep feature based duplication detection was developed. Images were firstly processed to feature vectors with a pre-trained deep model. In this case, we used a light-weight model, ResNet18 \citep{he2016deep}, which generates 512 dimensional feature vectors from resized input images of $224\times224\times3$. The assumption is then, that similar images should also be close to each other in feature space, which can be revealed using clustering algorithms.
%The duplication detection was mainly focused on the flood relevant images, where we applied a 50\% threshold to the probability outputs. A total of 4,601 images were used for duplication detection.
%
%We clustered the features using DBSCAN, a density-based clustering method. 
DBSCAN requires two parameters, \textit{eps} and \textit{minPts}, which represent the distance between the features in a cluster, and the required minimum number of elements in a cluster. \textit{minPts} in this case is 2 because we aim to include also the duplicated image pairs. A suitable \textit{eps} can be determined by a k-Nearest Neighbor Graph. As described in \cite{sander1998density}, by analyzing the sorted k-distances, good values are in the ``valley''. Different from most other applications of DBSCAN, in this case, the majority of images is considered as ``noise'' for DBSCAN, where the clusters of duplicated images are the minority.
Then, we selected the second significant turning point at 13 from the graphical representation shown in Figure \ref{fig:knn}. Among the retrieved 4,601 images, 207 clusters were identified. We manually checked the clusters which revealed that only three of the clusters contained non-duplicated images, whereas all of the remaining clusters indeed represented duplicated and near-duplicated images. To select the most relevant image for a cluster, the posted earliest image was preserved and all later ones were deleted. Overall, in this step, we were able to eliminate 653 duplicated images and finally ended up with 3,948 images for further processing.
Images from the largest cluster are shown in Figure \ref{fig:duplicate}, which cover different duplication cases, such as clipping, changing colour, and adding text. 

% OPTICS 763(totol duplicated) - 299(number of cluster)     464
% DBSCAN-13 860(totol duplicated) - 207(number of cluster)  653
% DBSCAN-12 714(totol duplicated) - 188(number of cluster)  526

\subsubsection{Water level estimation}

\iffalse
% Comparing different threshold splits
\begin{table*}[width=2\linewidth,pos=t]
	\caption{Evaluation of water level classifier considering different uncertainty scores}
	\begin{tabular*}{\tblwidth}{@{} LCCCCCC@{} }
		\toprule
		Threshold & Overall Accuracy & Macros Avg.Precision & Macros Avg.Recall & Macros Avg.\Fone-score & Support & Total Number\\
		\midrule
		Over 50\% & 74.45\% & 56.31\% & 68.83\% & 60.22\% & 1096 & 4601\\
		Over 80\% & 74.63\% & 57.75\% & 69.02\% & 61.16\% & 954 & 4129\\
		Over 95\% & 75.75\% & 60.57\% & 69.77\% & 63.52\% & 804 & 3644\\
		Over 99\% & \textbf{76.18\%} & \textbf{62.66\%} & \textbf{70.28\%} & \textbf{65.24\%} & 676 & 3142\\
		\bottomrule
	\end{tabular*}
	\label{tab:evaluation_wlevel_harvey}
\end{table*}
\fi

\begin{figure}
	\centering
	\includegraphics[clip=true, trim=55pt 10pt 35pt 25pt, width=.75\linewidth]{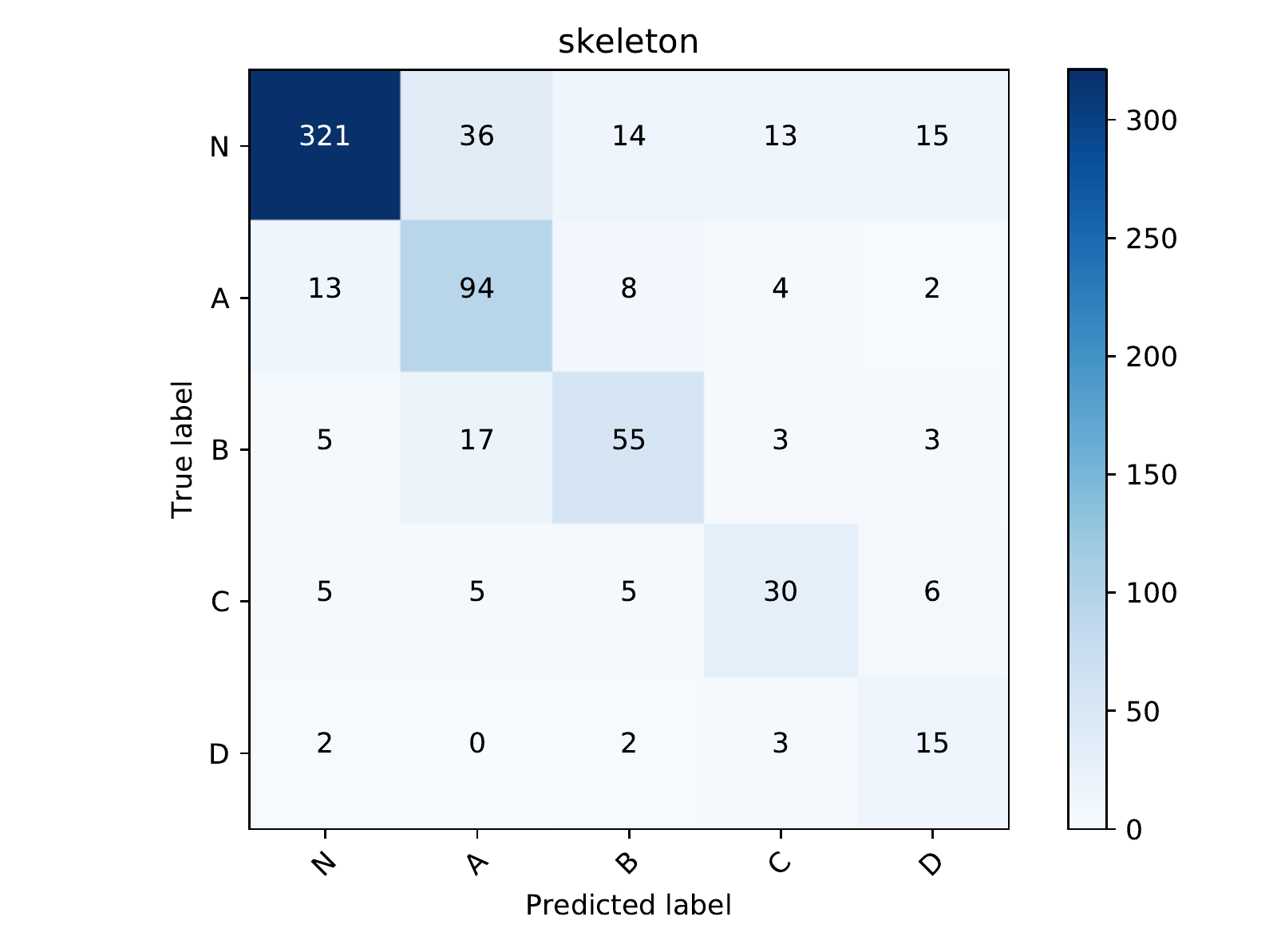}
	\caption{Confusion matrix of the water level estimation on social media images with flood relevance over 99\%}
	\label{fig:99evaluation_harvey}
\end{figure}

The resulting flood relevant images were further processed with the water level estimation model as described in Section \ref{sec:wlevel}. 
In this case, we only considered the images highly related to floods, i.e. with a confidence score over 99\%. 
%There are in total 3142 images.
After applying all the above described filtering processes, 676 flood-related images remained for the water level estimation. In order to evaluate the performance of our model for this real event, we annotated the images based on the annotation rules described in Section \ref{sec:water_level_dataset} and obtained the confusion matrix shown in Figure \ref{fig:99evaluation_harvey}. The overall accuracy of our model is 76.18\% with a 
macro averaged \Fone-score 65.24\%. %and 
%\hl{weighted averaged \Fone-score 77.18\%}. 
% micro average 76.18\%
The number of false positives and false negatives between the four water level classes are relatively small. However, there are many images, which are supposed to show no evidence for water level estimation (i.e. class \textit{N}), classified with a water level class.
Comparing these results with the ones in Section \ref{sec:methodology_experiments} (90\%), the reduction of performance may be due to two aspects. One is the image quality and type, e.g. the images in social media can have various sizes. The other is that the training examples of class \textit{N} can cover only a small fraction of the cases encountered in reality. 
Especially, people in the scenarios with other postures than standing have a higher chance to be wrongly predicted by the classifier (e.g. sitting as shown in Figure \ref{fig:failed}).

\subsection{Flood mapping from VGI}
\label{sec:flood_map}
After estimating water level from social media images containing people, the next step is to link these estimates to the locations on the map, with the goal of providing a map of the flood extent and a map of the flood severity. In order to evaluate our results, it is essential to have ground truth to compare. This is difficult, as an exact ground truth comparable to VGI is not available. The following data sets have been selected as reference: There is a data set with property damage claims from the U.S. Federal Emergency Management Administration (FEMA) \citep{fema2018damage}.
An additional data set -  Harvey flood depths grid dataset - contains modeled inundation from FEMA \citep{fema2018depth}. Furthermore, there is a map with flood extent marked by remote sensing detection from Dartmouth Flood Observatory \citep{dfo2017flood}. 
%A comparison of VGI-generated maps with these maps is difficult, as those data sets contain aggregated information over the whole disaster; VGI posts, however, have a highly temporal and local validity. Thus, the absence of information does not necessarily mean the absence of the phenomenon at that time and location; 
%however, the presence of a post is a relevant indicator xxxx  

In the following subsections, we present the mapping possibilities with the extracted information. First, the individual severity estimations including text and image are visualized as markers with pop-ups in Section \ref{sec:flood_points}. 
%This visualization can provide a straightforward overview on the spatial distribution of individual flood level related Tweets. 
%This map can provide detailed information about the flood severity at individual locations, together with an exact image. However, it does not provide an integrated overview of the flood extent and corresponding severity.
Flood extent was determined from VGI by aggregating the locations of flood related posts %according to the census tracts overlaid with flood related posts
(see Section \ref{sec:flood_extent}).
Flood severity was determined from VGI %for the census tracts containing 
by aggregating water level estimates (see Section \ref{sec:flood_severity}). 

\begin{figure*}
	\centering
	\includegraphics[width=1.5\columnwidth]{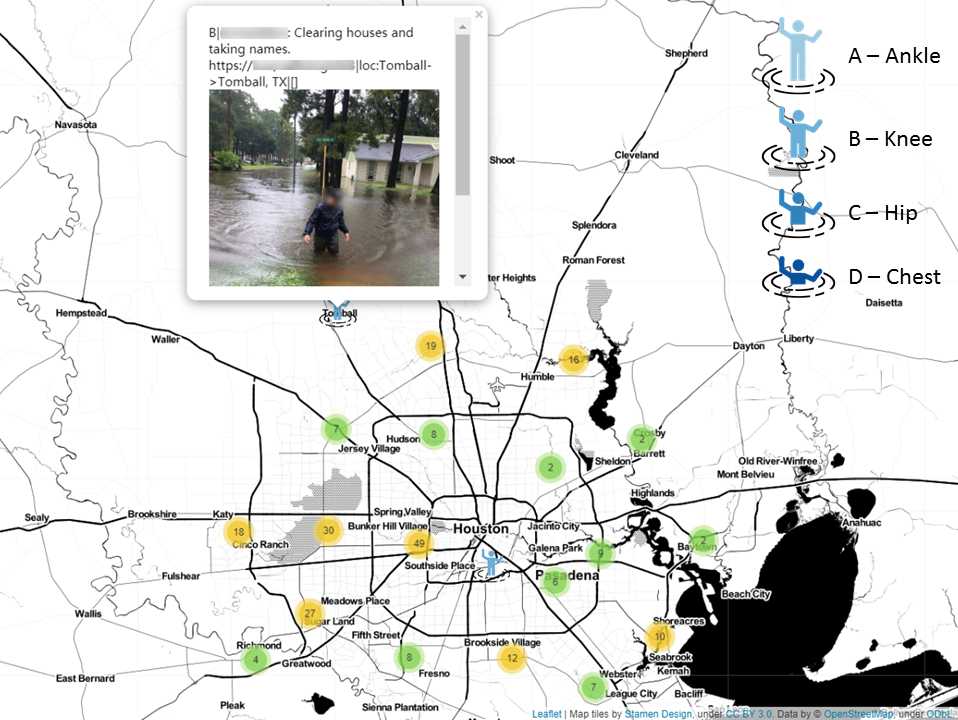}
	\caption{Map of social media posts with severity predictions as markers (Basemap: OpenStreetMap)} %\href{https://yuzzfeng.github.io/demos/20200220_Harvey_wlevel.html}{Demo}
	\label{fig:tweet_map}
\end{figure*}

\subsubsection{Map of individual severity estimation}
\label{sec:flood_points}

The locations of Tweets are generally given in three types. Type 1-Tweets provide exact geo-coordinates, which is a rare case, covering only 3.29\% of the total amount of our data. Type 2, 33.72\% of the retrieved Tweets, provide the location information corresponding to an area, where a bounding box is normally given. Type 3 (62.99\%) are the retrieved Tweets that are shared Instagram posts. 
Both geo-coordinates and bounding boxes are available. However, the saved geo-coordinates may represent either a point location or an administrative area such as city and district. Both cases are represented by point coordinates.
The recorded bounding box normally represents the corresponding city-level bounding-box.

The most straightforward way to present the extracted information on a web map is using markers with symbols representing the flood severity situation. In dense areas, markers are clustered. As shown in Figure \ref{fig:tweet_map}, users can click into the cluster to inspect individual Tweets on the web map. For types 1 and 3, we located the Tweets to the given coordinates. However, city-level Tweets do not provide much information about where the observations were taken. Therefore, we excluded the Tweets with city level geo-coordinates, such as for the City of Houston, or Harris County.
For type 2, where the Tweets have only bounding boxes, we positioned the Tweets at locations of box centres. 

This visualization can provide a straightforward over\-view on the spatial distribution of individual flood level related Tweets. This map can provide detailed information about the flood severity at individual locations, together with an exact image. However, it does not provide an integrated overview of the flood extent and corresponding severity.

%This map can provide detailed information about the flood severity at individual locations, together with an exact image, however, \hl{it does not provide an integrated overview of the flood extent and corresponding severity.}
%\hl{cannot} provide a straightforward view of flood extent and corresponding severity.

\subsubsection{Map of flood extent}
\label{sec:flood_extent}

\begin{figure*}
	\centering
	\includegraphics[clip=true,trim=0pt 0pt 0pt 0pt, width=0.9\linewidth]{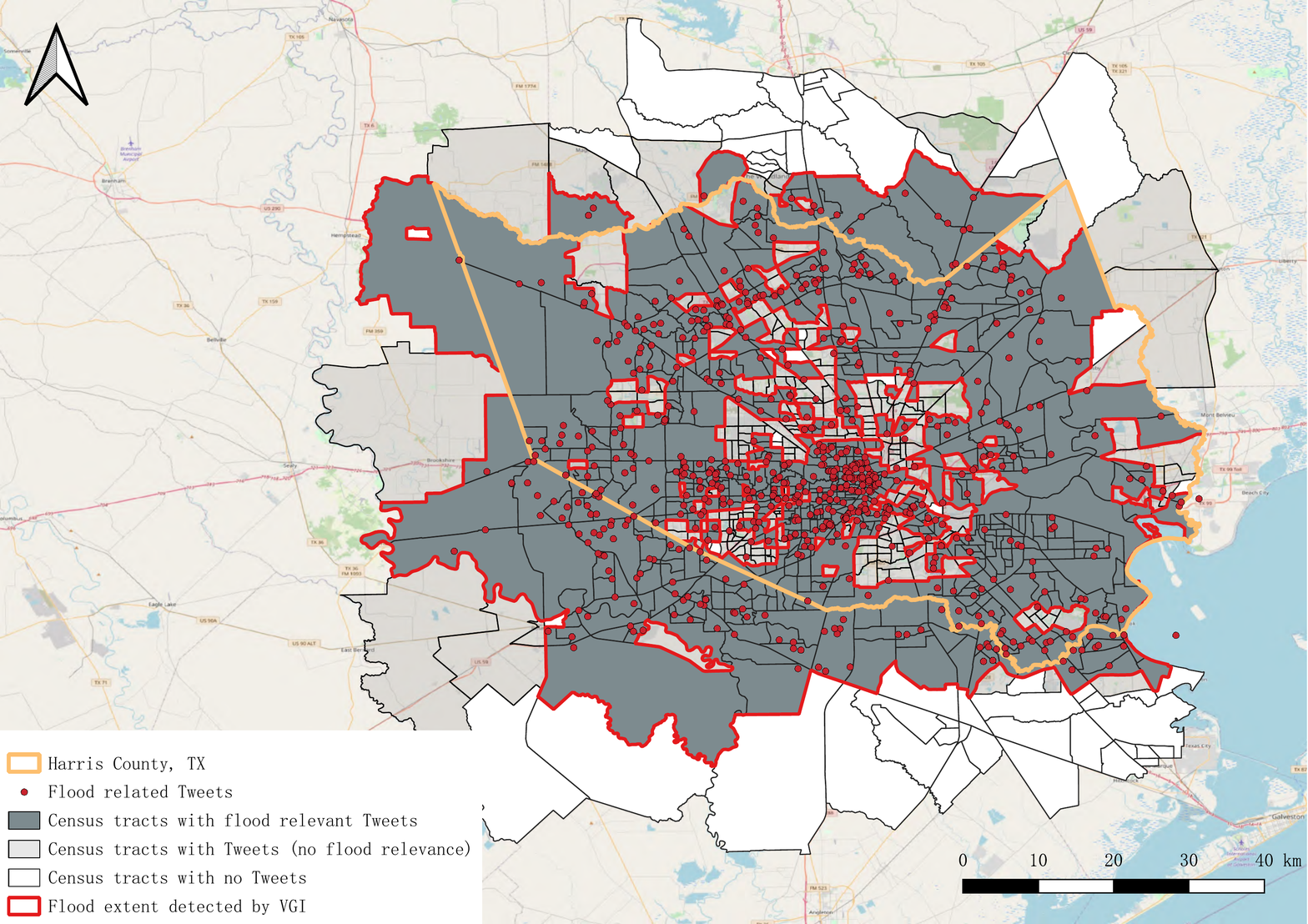}
	\caption{Locations of flood relevant Tweets with overlaid census tracts as the flood extent detected by VGI}
	\label{fig:vgi_map}
%\end{figure*}
%\begin{figure*}
	\centering
\includegraphics[clip=true,trim=0pt 0pt 0pt 0pt, width=0.9\linewidth]{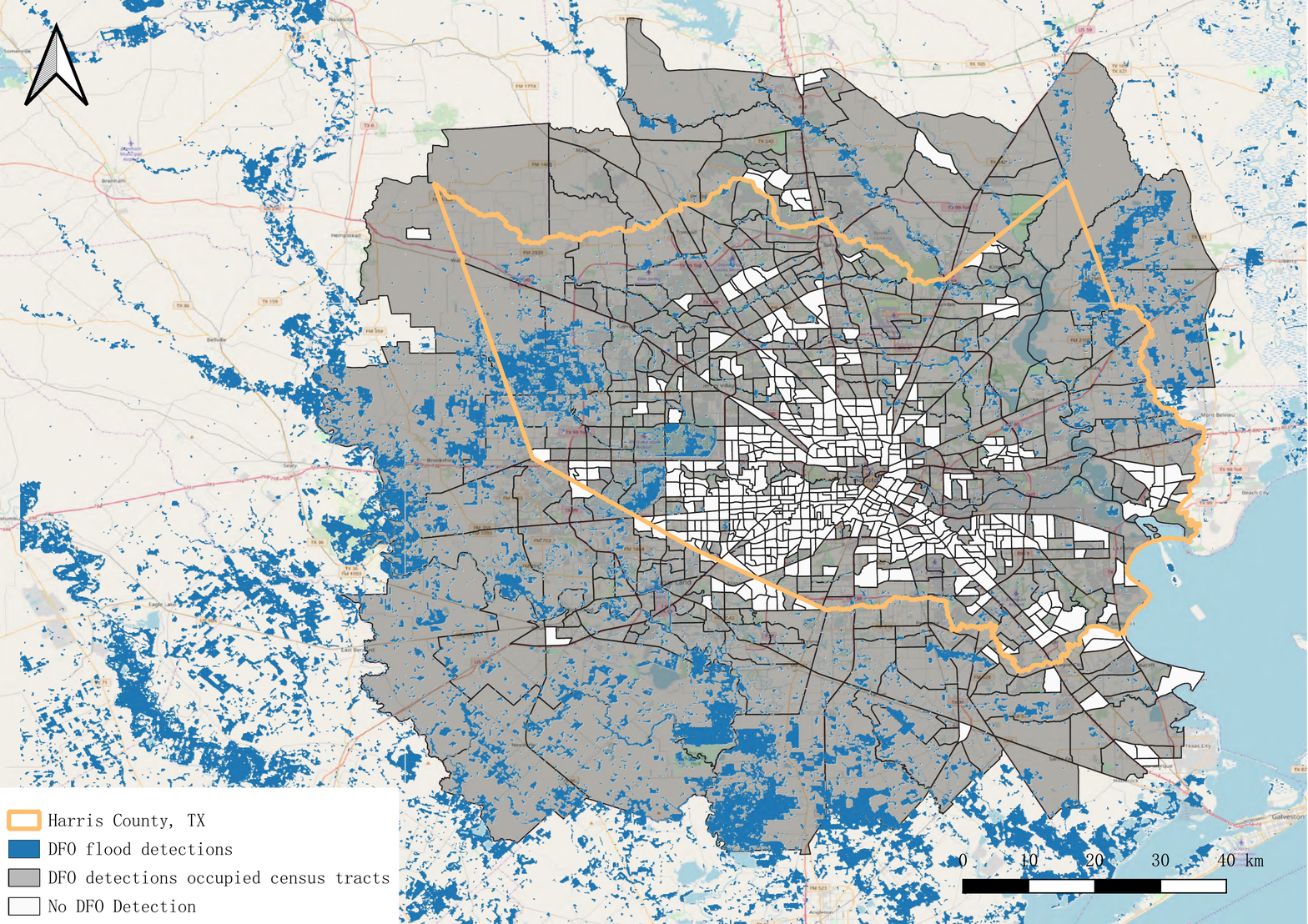}
	\caption{Maximum observed flooding mapped from NASA MODIS, ESA Sentinel 1, ASI COSMO-SkyMed, and RADARSAT 2 data from Dartmouth Flood Observatory~\citep{dfo2017flood} and the overlaid census tracts}
	\label{fig:rs_map}
\end{figure*}

\begin{figure*}
	\centering
	\includegraphics[clip=true,trim=0pt 0pt 0pt 0pt, width=0.9\linewidth]{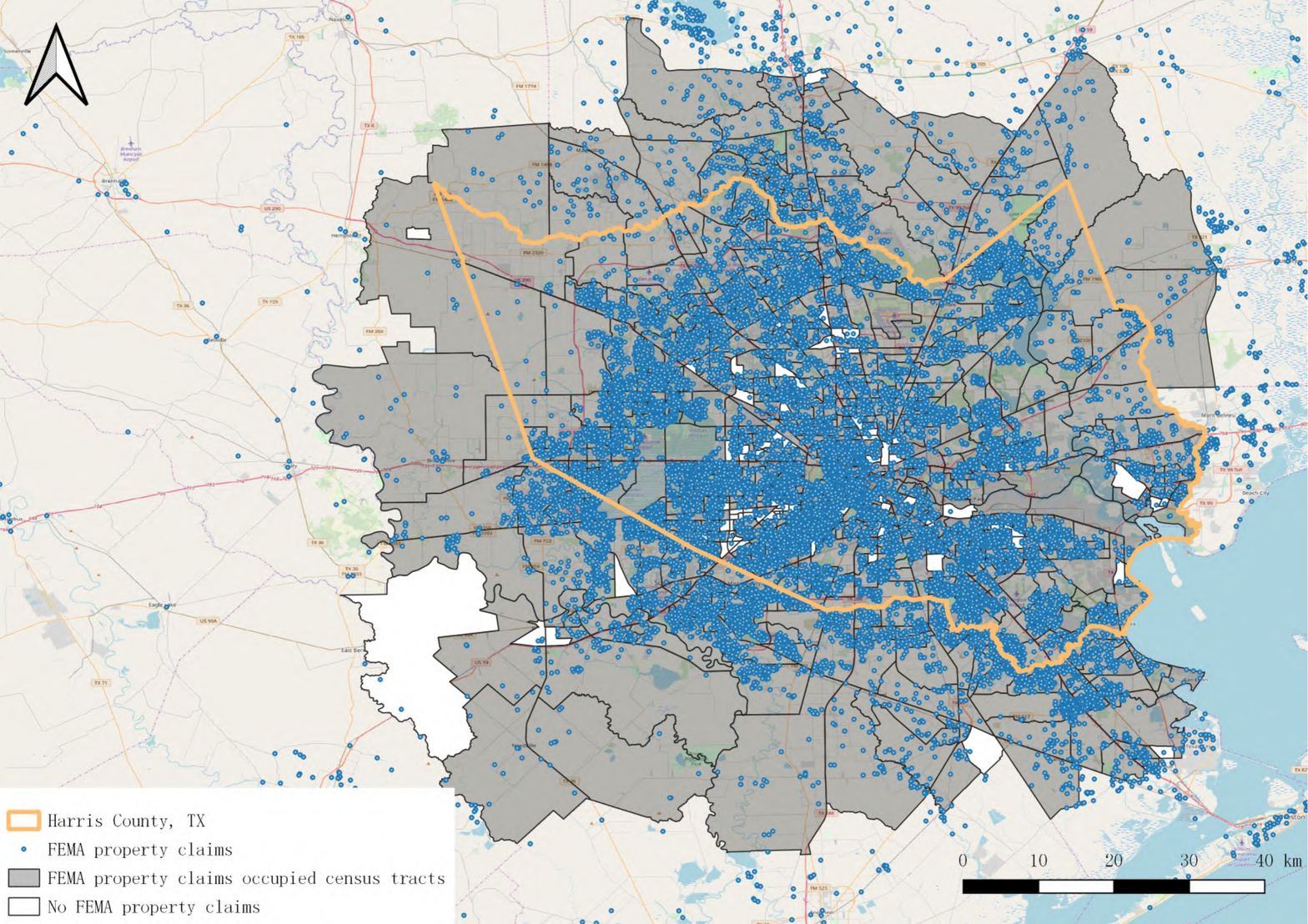}
	\caption{FEMA property claims and the overlaid census tracts. Data source: \cite{fema2018damage}}
	\label{fig:fema_map}
%\end{figure*}
%\begin{figure*}
	\centering
	\includegraphics[clip=true,trim=0pt 0pt 0pt 0pt, width=0.9\linewidth]{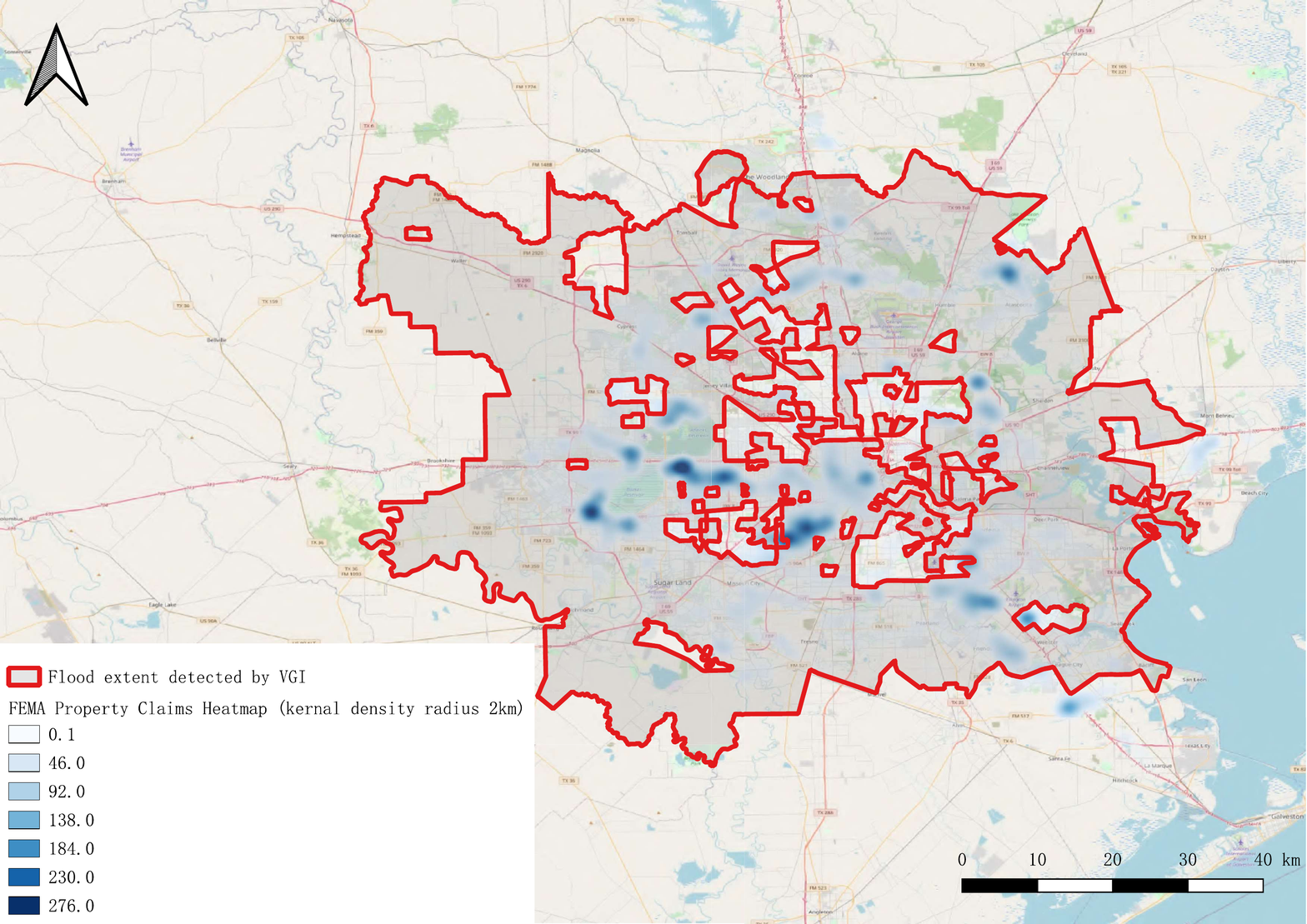}
	\caption{FEMA property claims density map and the flood extent detected by VGI. Data source: \cite{fema2018damage}}
	\label{fig:fema_vgi_map}
\end{figure*}

In order to achieve an overview of the flood situation, the point information given with the Tweets have to be extended to areal information, typically using spatial interpolation methods. However, in our case, social media posts are very sparsely and unevenly distributed in space. The main factor for inundation - terrain - varies from regions to regions significantly. Thus, interpolation can hardly reflect the real situation between observations. Additionally, the locations of the Tweets may refer to either a point location or a bounding box.
Therefore, instead of interpolation, aggregation of the information to spatial units is a more reasonable representation for the flood situation.

In the United States, the most commonly used spatial units in geography are census tracts. They are relatively permanent statistical subdivisions of a county, which have on average about 4,000 inhabitants \citep{us2015census}. We downloaded the boundary files for Texas available at \cite{us2018mapping} and extracted the tracts around Harris County, which covered most of the Houston metropolitan area. This area contains 966 census tracts in total.
%metro\-po\-li\-tan solved by import sloppy 

%% Delete because the contents in 4.2 moved here
%As described in Section~\ref{sec:flood_map}, aggregation of the extracted information into spatial units of the census tracts was selected to present flood extent.
From the Tweets collected within 8 days of Hurricane Harvey, we marked the census tracts where Tweets were sent with light grey colour, and the census tracts where flood relevant Tweets were sent with dark grey colour, shown in Figure \ref{fig:vgi_map}. For Tweets with exact coordinates (type 1 and 3), in order to consider probable location uncertainty, the tracts overlaid with a buffer around the coordinate were marked. As described in an analysis on the positional accuracy of social media images \citep{cvetojevic2016positional}, the distances between the image content and photo upload location have a median value of 198.7m for Twitter in North America and 85m for Instagram posts. Therefore, we considered a 200m radius for the location of social media post to eliminate the potential issue where posts located very close to the tract border.
For the Tweets only with bounding boxes (type 2), all the intersected tracts were marked. The area (with holes) marked with a red boundary in Figure \ref{fig:vgi_map} is the flood extent estimated by VGI.

% Considered the border effect in 1.revision
\begin{table}[width=.9\linewidth,pos=t]
\caption{Comparison of water extent mapping from different information sources}
\begin{tabular*}{\tblwidth}{@{} CCCCC@{} }
\toprule
Method & Precision & Recall & \Fone-score & Accuracy\\
\midrule
VGI & 96.54\% & 64.68\%  & 77.46\% & 64.70\%\\
RS & \textbf{97.10\%} & 51.77\%  & 67.53\% & 53.31\%\\
VGI+RS & 96.35\% & \textbf{81.68\%}  & \textbf{88.41\%} & \textbf{79.92\%}\\
\bottomrule
\end{tabular*} 
\label{tab:compare_vgi_rs}
%\end{table} 
%\begin{table}[width=.9\linewidth]
\caption{Confusion matrices of water extent mapping from different information sources}
\begin{tabular*}{\tblwidth}{@{} CCccCccCcc@{} }
\toprule
Method &&\multicolumn{2}{c}{VGI} && \multicolumn{2}{c}{RS} && \multicolumn{2}{c}{VGI+RS} \\
\cmidrule(l){1-1} \cmidrule(l){3-4} \cmidrule(l){6-7} \cmidrule(l){9-10}
Predicted && 0 & 1 && 0 & 1 && 0 & 1\\
\midrule
True & 0 & 39 & 21  && 46 & 14 && 32 & 28 \\
Labels & 1 & 320 & 586  && 437 & 469 && 166 & 740\\
\bottomrule
\end{tabular*} 
\label{tab:compare_vgi_rs_cm}
\end{table} 

\iffalse % 1.Submission
\begin{table}[width=.9\linewidth,pos=t]
\caption{Comparison of water extent mapping from different information sources}
\begin{tabular*}{\tblwidth}{@{} CCCCC@{} }
\toprule
Method & Precision & Recall & \Fone-score & Accuracy\\
\midrule
VGI & 96.40\% & 62.03\%  & 75.49\% & 62.22\%\\
RS & \textbf{97.10\%} & 51.77\%  & 67.53\% & 53.31\%\\
VGI+RS & 96.38\% & \textbf{79.36\%}  & \textbf{87.05\%} & \textbf{77.85\%}\\
\bottomrule
\end{tabular*} 
\label{tab:compare_vgi_rs}
%\end{table} 
%\begin{table}[width=.9\linewidth]
\caption{\hl{Confusion matrix of water extent mapping from different information sources}}
\begin{tabular*}{\tblwidth}{@{} CCCCCCCCCC@{} }
\toprule
Method &&\multicolumn{2}{c}{VGI} && \multicolumn{2}{c}{RS} && \multicolumn{2}{c}{VGI+RS} \\
Predicted && 0 & 1 && 0 & 1 && 0 & 1\\
\midrule
True & 0 & 39 & 21  && 46 & 14 && 33 & 27 \\
Labels & 1 & 344 & 562  && 437 & 469 && 187 & 719\\
\bottomrule
\end{tabular*} 
\label{tab:compare_vgi_rs_cm}
\end{table} 
\fi

%\hl{In order to evaluate the performance of the flood extent mapping, the FEMA property damage claims were used as the reference. Census tract level flood extent from Dartmouth Flood Observatory was used as baseline. Subsequently, we compared the VGI detected flood extent, remote sensing detected flood extent, and the fused flood extent of both detection at census tract level.}

Remote sensing has been widely applied for flood extent mapping and used in this research as a baseline.
The remote sensing detection is from the Dartmouth Flood Observatory. They extracted the maximum observed flooding for Hurricane Harvey from NASA MODIS, ESA Sentinel 1, ASI Cosmo SkyMed, and Radarsat 2 data \citep{dfo2017flood}, shown as blue pixels (of size 85 m $\times$ 85 m) in Figure \ref{fig:rs_map}. We aggregated the pixels to census tracts by overlay and marked them in grey. It can be observed that there is less flooding observed in the city center, whereas more flood pixels are detected outside the city or along the river in the city. 

The property claims for hurricane Harvey from \cite{fema2018damage} was used as a reference. It contains property damage claim with dates, loss types (e.g. electric current, wind, flood, water damage), and locations (both in text and coordinate). From 27\textsuperscript{th} of August to 2\textsuperscript{nd} of September 2017, in total, 226,167 property claims were collected in Texas and 38,422 of them are caused by flood or water damage. 
Since this data was collected for insurance purpose, non-critical errors are allowed \citep{fema2019national}. In order to aggregate the points to census tracts considering this probable uncertainty, only the tracts with 3 or more claims were considered as flooded regions. This threshold was chosen based on the distribution of the number of claims in each cell. Since this distribution is skewed, log transformation was applied to the data to achieve a distribution similar to a normal distribution. Afterwards, a confidence interval of 2-sigma was applied to this transformed distribution. The cells with a transformed score smaller than this interval were regarded as outliers, which correspond to the cells with 1 or 2 claims in the original distribution.
As shown in Figure \ref{fig:fema_map}, since Hurricane Harvey is a great disaster which led to huge lost, most of the tracts contains property damage claims caused by flood or water.

%It can efficiently detect changes of water body over times, which provides precise flood extent mapping. However, for urban flooding, the resolution of the satellite image products influence the ability to capture details of complex urban landscapes. However, high-resolution images (e.g. Ikonos, QuickBird) provide detailed observation than the relatively lower-resolution images (e.g. TM, ENVISAT), but the long revisit circle and exorbitant prices limits the data availability for urban flood monitoring. \citep{feng2015urban}
% In this work, we \hl{select} the property claims where people \hl{reported} their property damage for insurance aftermath. 
%Since the claims may contain uncertainty and no detailed information such as level of damage is available, we aggregated the points to census tracts with a count threshold 3. 

We then treated the FEMA property claims as ground truth and compared them with the detected flooded tracts from both remote sensing and VGI. We also merged both results by the logic OR operation. The precision, recall, \Fone-score of the positive class and overall accuracy at census tract level are summarized in Table \ref{tab:compare_vgi_rs} and the confusion matrices are summarized in Table \ref{tab:compare_vgi_rs_cm}.

%Number of claims in all cells 28612
%Number of claims in extent 22740
%79.48% claims are included within the estimated flood extent.

\begin{figure*}
	\centering
	\begin{minipage}{0.495\linewidth}
		\centerline{\includegraphics[clip=true,trim=0pt 0pt 0pt 0pt, width=\linewidth]{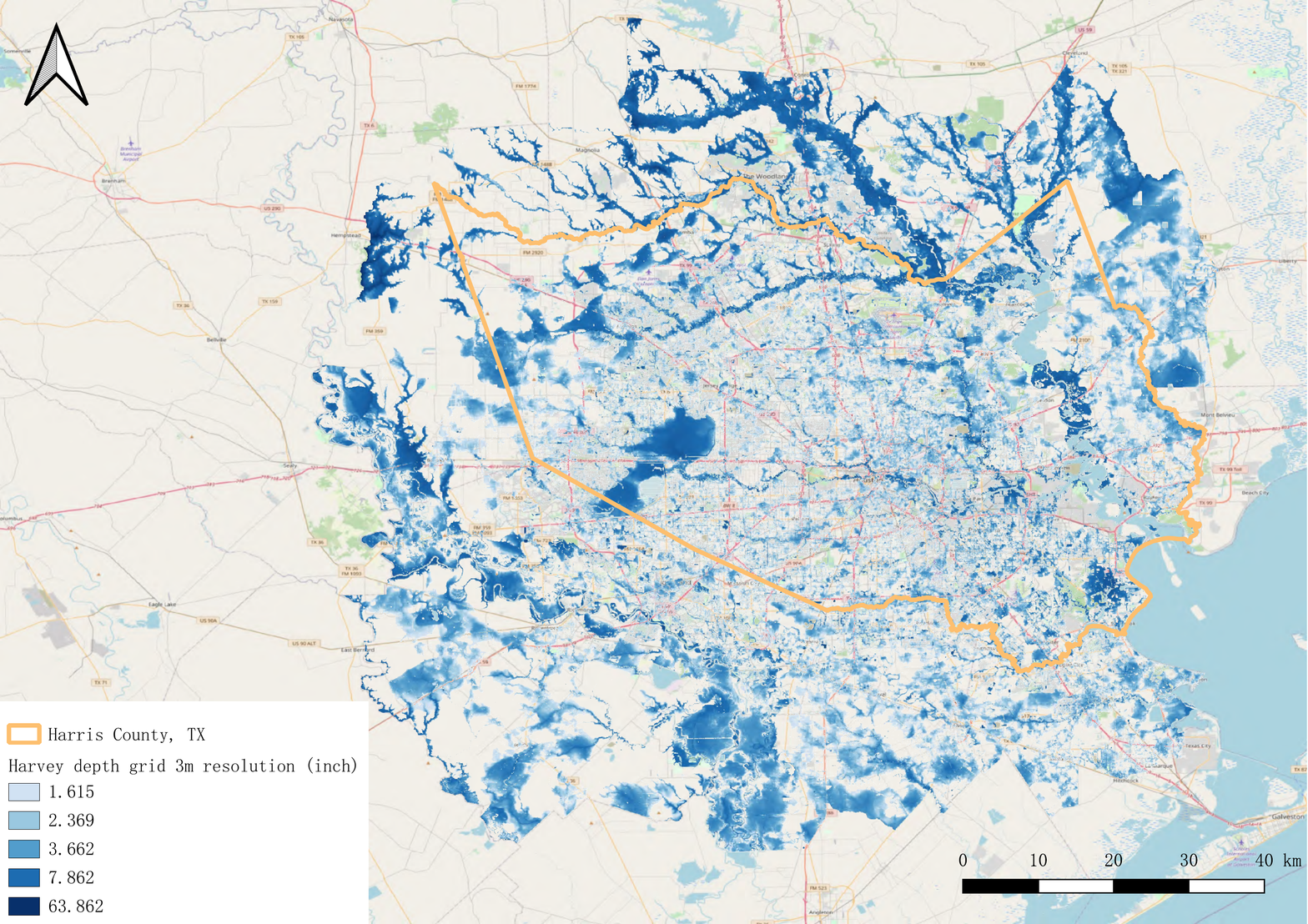}}
		\caption{FEMA Harvey flood depth grid \citep{fema2018depth}}
		\label{fig:depth_grid}
	\end{minipage}
	\hfill
	\begin{minipage}{0.495\linewidth}
		\centerline{\includegraphics[clip=true,trim=0pt 0pt 0pt 0pt, width=\linewidth]{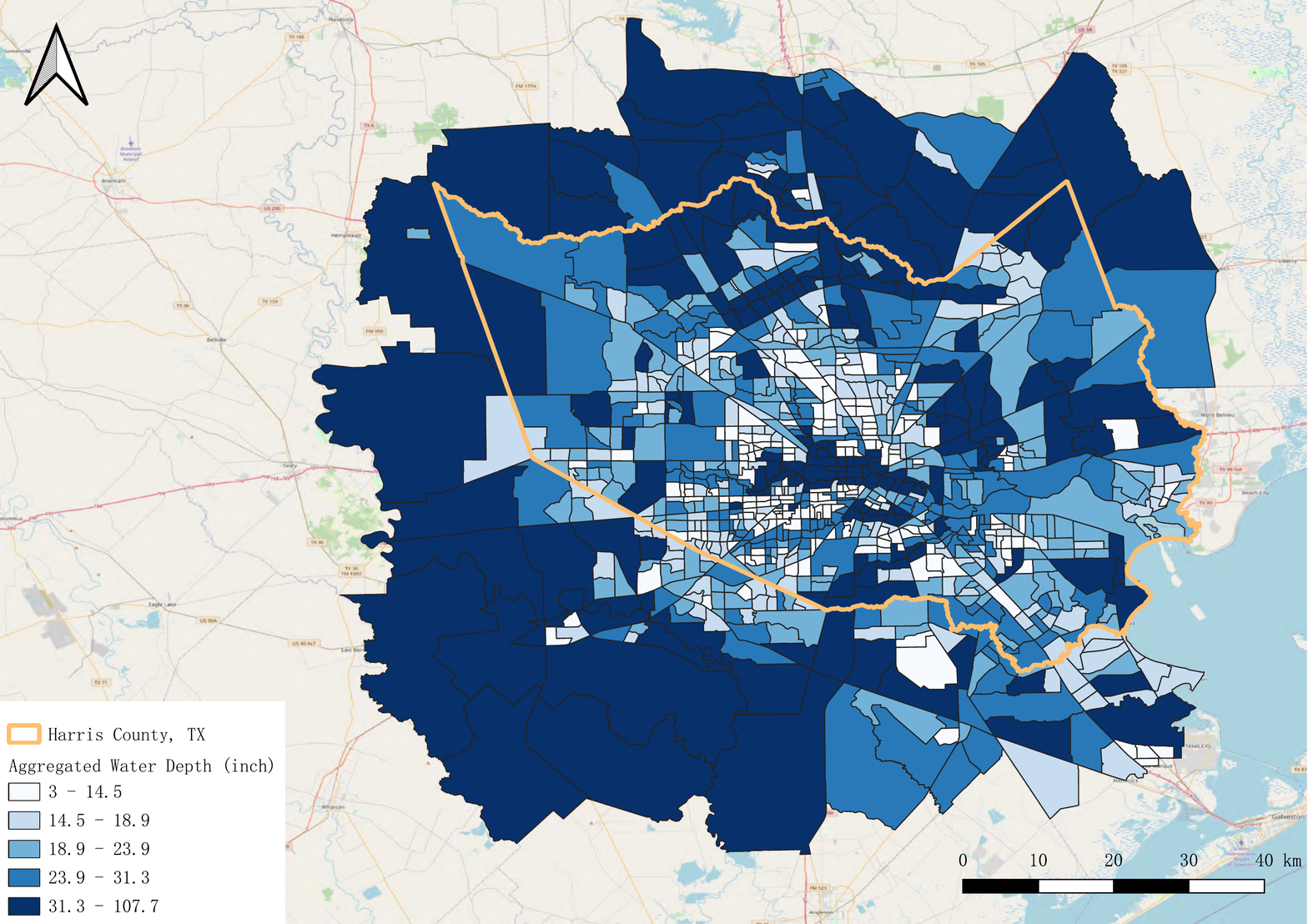}}
		\caption{Aggregated flood depth tracts with max depth values}
		\label{fig:depth_tracts}
	\end{minipage}
\end{figure*}		
\begin{figure*}
	\centering
	\includegraphics[clip=true,trim=0pt 0pt 0pt 0pt, width=0.9\linewidth]{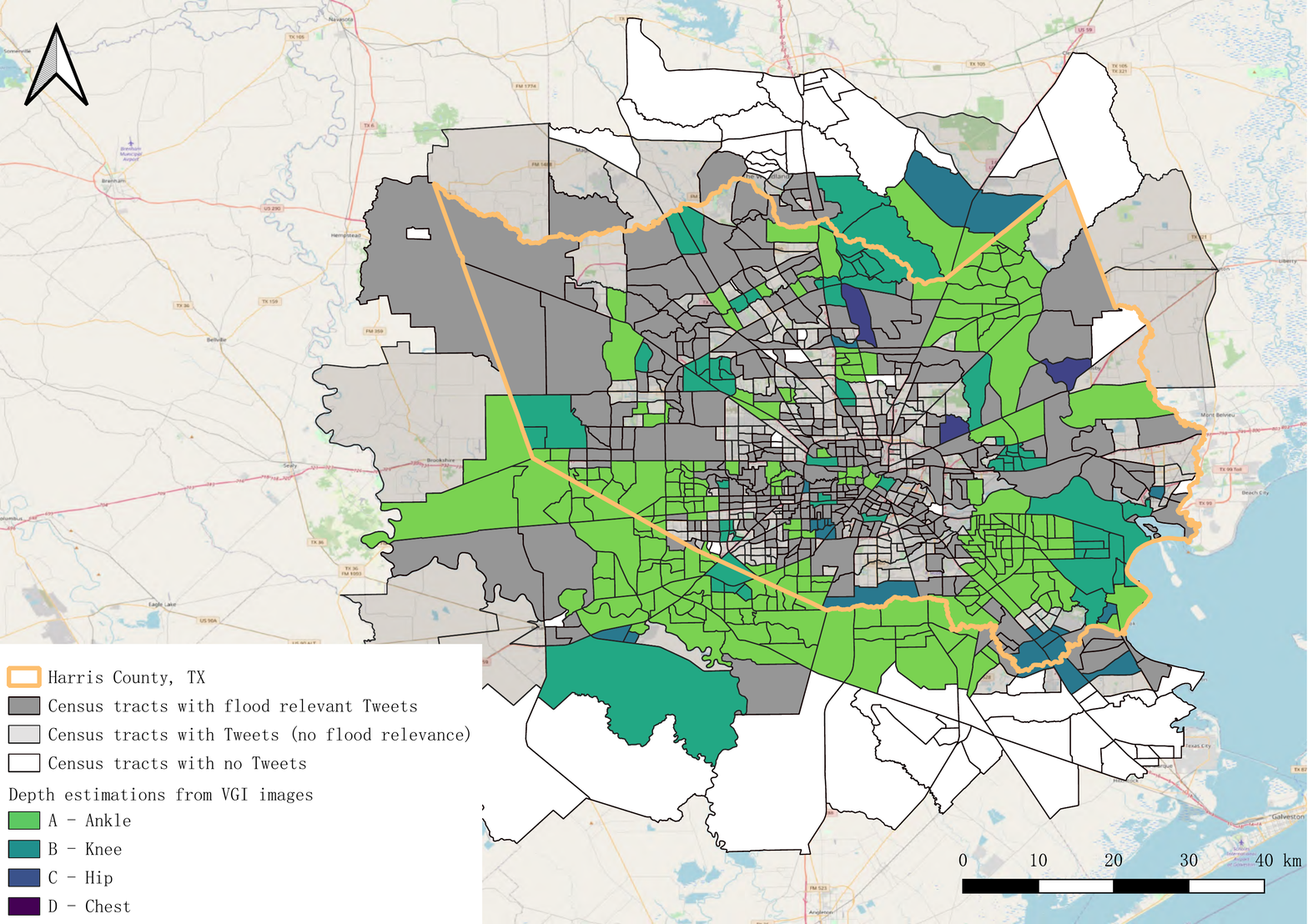}
	\caption{Flood severity map derived from the water level estimations of VGI images}
	\label{fig:wlevel_map}
\end{figure*}

According to Table \ref{tab:compare_vgi_rs}, remote sensing detection achieved the best precision but also a low recall. Based on a visual comparison between the remote sensing (Figure \ref{fig:rs_map}) detection and the reference (Figure \ref{fig:fema_map}), many false negatives are located in the city centres. Even though VGI provided only very sparse spatially distributed data points, it was able to mark the flooded census tracts with only a slightly lower precision but a higher recall compared to the remote sensing detection. Based on a visual comparison between the VGI based detection and the reference, more census tracts in the city centre are correctly detected. However, due to the lack of observations in census tracts where no Tweets are available (as the white tracts shown in Figure \ref{fig:vgi_map}), there are still many false negatives which lead to a low recall of 64.7\%.
Simply combining the VGI and remote sensing detection achieves a much better overall accuracy and \Fone-score, which shows the complementary properties of VGI. With this, we proved that VGI can be used as a supplement data source for flood extent mapping, especially beneficial for urban areas.

To regionalize the information, we applied kernel density estimation with a radius of 2 km on the FEMA property claims and overlaid it with the area VGI marked as flood extent in Figure \ref{fig:fema_vgi_map}. It was identified that almost all the ``heat regions'' are located within the red border of the flood extent marked by VGI, especially the two heat regions in the west and southwest of the city center. 
Our flood extent from VGI excludes the census tracts where there are no significant heat regions in the city's northeast, northwest and southeast. Users in most of these areas sent Tweets, but no images related to the flood appeared.

\subsubsection{Map of flood severity}
\label{sec:flood_severity}

% [CB] sect. 4.2 can't be referenced here since we are still in sect. 4.2. Or should this link to 4.2.2.? 
The Harvey flood depth grid dataset was used as the reference to evaluate the performance of flood severity mapping. It is in 3~m resolution and was published by FEMA on 15\textsuperscript{th} of November 2017 \citep{fema2018depth}. It was generated based on High Water Marks from aftermath field survey and Digital Terrain Models in the form of a Triangulated Irregular Network (TIN). Four quality assurance measures (namely identifying dips, spikes, duplication, and inaccurate/unrealistic measurements) were applied. In addition, the water areas were removed based on authoritative data \citep{census2019tiger}.
The flood depth data in our study area are visualized in Figure \ref{fig:depth_grid}.
Since our severity estimation from VGI is at census tract level, the water depths were aggregated to census tracts by calculating the maximum flood depth to represent the most severe situation of each census tract (shown in Figure \ref{fig:depth_tracts}).

%\hl{In order to evaluate the performance of our flood severity mapping, the Harvey flood depths grid dataset by FEMA was used as the reference.}
%\hl{Subsequently, the correlation was calculated between the estimated water level class and the modeled water depth at census tract level.}
Among the 966 census tracts observed in this study, 312 of them could provide flood severity estimations based on the interpretation of social media images. Flood severity estimations were aggregated to tracts (shown in Figure \ref{fig:wlevel_map}) according to the most frequent flood severity class.
Subsequently, we calculated the correlation between VGI estimated flood severity and water depth from FEMA to evaluate the performance of VGI based flood severity mapping.
Since the VGI based flood severity estimations are ordinal and skewed while the modeled water depths are continuous and skewed, Spearman's rank correlation is an appropriate correlation coefficient to use according to \cite{mukaka2012guide}.
The result 
%($r=0.2042$, $n=312$, $p<0.001$)  % 1.Submission
($r=0.1836$, $n=323$, $p<0.001$)  % 1.Revision considered border effect
indicates a weak positive monotonic correlation with high significance between these two variables. This is based on the interpretation for positive correlation (weak: r>0.1, moderate: r>0.4, strong: r>0.7 and perfect: r=1) in \cite{akoglu2018user}.

Due to the sparse and uneven distribution of VGI, the number of the VGI data points available in each census tract is sometimes very limited. 140 out 312 tracts have only 1 or 2 valid images for severity mapping. Nevertheless, these real-time observations can already provide an integrated overview of flood severity which has a weak positive association to the real situation.
It is also worth noting that, even though this information is few and sparse, it is normally available well in advance of the remote sensing observations, which is valuable during the emergency response phase.

%By inspecting in combination with the individual water level estimation markers as presented in Section \ref{sec:flood_points}, decision-makers can get an intuitive situation awareness, where severe inundation happens and how severe the situation is at the very moment. It is worth noting that, even though this information is few and sparse, it is normally available well in advance of the remote sensing observations, which is valuable during the emergency response phase. In addition, the automatic flood severity interpretation from images extends the usefulness of VGI and provides %platform 
%\hl{users} with the most evident information efficiently.

%bitte prüfen: manche der aussagen in dem obigen abschnitt sind allgemeiner natur und könnten auch in die discussion section; dort haben sie bislang hauptsächlich die limitations aufgezählt - ev. könnte was von den oben genannten vorteilen auch noch dort präsentiert werden. 

%In another research, where water level information were manually interpreted, they conclude there are a statistically significant difference between interpolated VGI and FEMA depth grids.
%Dangerous area when people cannot stand cannot be observed by our process. Also social media users do not necessarily got the chance to observe the most severe situation in their tracts.
% current - max wdepth \hl{(r=.2042, n=312, p<.001)}
% old version - median wdepth \hl{(r=.1716, n=312, p=.002)}

\section{Discussion}
%Authors should discuss the results and how they can be interpreted in perspective of previous studies and of the working hypotheses. The findings and their implications should be discussed in the broadest context possible. Future research directions may also be highlighted.

As presented in the sections above, we have proposed a process, bundling different methods to collect, retrieve and analyze social media images of flood events. The different elements in the process successfully extracted flood relevant information, removed duplicates and classified the water level.

Flood severity information was extracted by analyzing these user-uploaded images, which can be a new information source for the city emergency response.
As suggested in the previous studies, we confirmed that objects with approximately known dimensions and submerged in water are good indicators for water level estimation.
We targeted on the people in the scene, where component level information (i.e. human pose) was used to support the water level classification. Compared with the baselines using deep features of the whole image and deep features around the detected people, our proposed method achieved better performance. The component level information is thus proved to be beneficial and inspire us to include the component level information of other potential objects (e.g. vehicle, bicycle) in our future work.
Furthermore, the annotation effort is significantly reduced, where each photo is annotated with only a single water level label instead of a time-consuming pixel level annotation.
%Fine-grained classification of more than two classes can be achieved, and our method is also easily adaptable for finer water level classes when the corresponding annotations are available.

This technical process was applied to a real event, Hurricane Harvey. 
The high accuracy of the benchmark experiments could not be achieved - this is attributed to the fact that in the Harvey scenario, many Twitter and Instagram photos were used, which are of lower quality than the training data.
Social media users may often overlay extra texts on photos, make collages from several photos, and also upload blurred photos in bad light condition. These unexpected situations are challenging not only for our proposed water level estimation method but also for most of the computer vision algorithms to detect objects, segment images, or extract human poses.

Due to differences in people's size and unknown perspectives, automatic interpretation of water levels from social media images can hardly reach centimeter-level accuracy as water engineering needs. 
However, it is more practical to use it in applications with lower accuracy requirements, e.g. emergency response or to improve the situation awareness of residents. The water level extracted from social media images is an intuitive indicator for flood severity. Rescuers and citizens would not take actions according to the extracted information alone, but also combine with their own interpretation on the information.

%However, our process can not only provide them with flood-related images, but also provide them with severity levels.
%however, most of the off-topic posts and posts that show no evidence on flood level are eliminated by our process. It can assist experts as involved in \cite{fohringer2015social} to further achieve cm-level water level annotation. In practise, this process is is more suitable for the application with lower requirement on accuracy, where the severity information is needed but do have to be precise.
%wrong predictions caused by wide variety of social media image or the image contents irrelevant to the event and location can be quickly discovered. However, these are nowadays still very challenging for deep learning algorithms and need further investigation.

Since remote sensing data used for disaster monitoring usually has a time delay, the information extracted from VGI can provide city managers with information on flood extent and severity at an earlier point in time. 
As presented in Section \ref{sec:flood_extent}, VGI can provide more observations for populated areas, whereas remote sensing is good at detecting flood water in less constructed area. Therefore, VGI can be used as a good supplement to remote sensing flood detection and delineation. 

Even though the extract flood severity map demonstrates only a weak correlation to the modeled results, by inspecting in combination with the individual water level estimation markers as presented in Section \ref{sec:flood_points}, decision-makers can get an intuitive situation awareness, where severe inundation happens and how severe the situation is at the very moment.
In addition, the automatic flood severity interpretation from images extends the usefulness of VGI and provides users with the most evident information efficiently.

%It is worth noting that, even though this information is few and sparse, it is normally available well in advance of the remote sensing observations, which is valuable during the emergency response phase. 

However, from the perspective of using social media as a source of information for flood monitoring, several limitations exist, which are discussed in the following: VGI extracted from social media mainly includes three parts, time, location and semantics. Each part may introduce uncertainty to flood mapping. 

In terms of time, there is typically a delay between when the user observed the event and when the photo was uploaded. This delay ranges from seconds to days, and varies from individual to individual, which can hardly be detected or quantified. 

In terms of location, social media users may send their posts with a fake or inaccurate location. Two common situations were observed. One is that people share information from other users' observations or news media images at their current location. Many of these cases can be detected by our image duplication detector based on the assumption that their shared images are the same or similar. The other situation is that people assign a wrong location intentionally or unintentionally. This case cannot be easily solved by the interpretation of the images. 
The location inference from images has been studied by other researchers, however, the position accuracy is insufficient for a city level flood mapping. 
Nevertheless, when detailed location information was mentioned by the user in the text, the Named-Entity Recognition (NER) based Geoparser \citep[e.g.][]{wang2019dm_nlp} has demonstrated a great potential to provide these images with a more precise geolocation. In this way, some of the user-generated locations can be verified and more images without location information can be used for flood severity mapping.
%and the availability of our \hl{process} can also be extended due to more flood severity observations.

In terms of semantic information, especially for extracting flood relevant information from images, photo editing and low-quality images are great challenges in many cases. 
In general, these problems can be mitigated, when several posts at a certain location and time are available; then it is possible to determine the earliest one - still it does not guarantee to indicate the exact time of the event. Also a majority filter concerning the semantics can be applied.

Even if VGI data are sparse and are provided with varying intensity in space and time (and quality), also several interesting inferences can be drawn: if there is an evidence, it also refers to an event; if there are many tweets in the region, but no flood-relevant tweets, then there is a high probability that there is no flood event. 

Concerning the water level estimation, we can identify three limitations of our current method.
First, our model cannot consider the difference in height between individuals, nor the difference between adults and children. The children in the scene may cause an overestimation of the flood severity.
Second, there are posts containing multiple images with different water level at the same location. We currently used the voting strategy to aggregate the multiple flood severity estimations to the location on the map. 
In order to solve this problem, additional information from the scene has to be taken into account. This requires a scene localization process \citep[e.g.][]{cattaneo2019cmrnet}.
%Precise difference in location cannot be considered even though multiple images are available.
Third, the number of images that can be used for water level estimation is limited. In this work, 676 out of 3,142 flood relevant images could be used for water level mapping during Hurricane Harvey. The number of useful images might be much fewer for a less significant event or events in less populated areas. The only fixed-size objects used for water level estimation are the persons in the scenarios. Therefore, other fixed-size objects could be introduced to overcome this issue, such as cars or bikes. 

For most VGI-based applications, data quality is a great challenge. Thus it can be beneficial to use a visual analytics approach, with the human in the loop. 
Since our method has eliminated most off-topic Tweets and Tweets that show no evidence on flooding, users need much less effort to verify model predictions and improve location quality. As for the 20,824 images during the 8 days of Hurricane Harvey, only 330 images with water level prediction need to be validated for correctness.

Social media is a real-time data source. With the models trained in advance, this real-time property can be preserved by setting up a proper infrastructure to analyze the data. OpenPose and Mask R-CNN have been proved to achieve a real-time performance in \cite{cao2018openpose, he2017mask}. Nowadays, there are also emerging solutions to achieve a real-time performance on semantic segmentation \citep[e.g.][]{yu2018bisenet}. Xgboost used in our methods for classification can also be deployed as a real-time online service \citep{google2018serving}. Nevertheless, a systematic time budget calculation is still needed, however, it is beyond the scope of our current work.

%\hl{As for the flood extent mapping at the very end of the process, in this work, a census tract level mapping result has been presented. However, by combining with DEM and hydraulic models, there is a great chance to achieve fine-grained flood mapping as in} \cite{eilander2016harvesting}. \hl{However, our current water level classification output is in discrete water level categories instead of numeric values. Therefore, further investigation is still needed to integrate such discrete information while considering the uncertainty of the water level estimation.} 

\section{Conclusions}
\label{sec:conclusions}

In this paper, we presented a novel process for mapping flood severity from social media images with location and applied it to a real flood event as a proof of concept. 
The process includes the collection and filtering of social media images with respect to flood relevant eyewitness pictures, as well as elimination of similar (and thus potentially duplicated) images. 
Furthermore, the flood relevant images containing people were classified into four flood severity levels according to the water level with respect to different body parts of people present in the scene. The water level estimation on a representative data set achieved an accuracy of 90\%. 
Compared with previous studies, our model achieved fine-grained water level classification with less annotation effort.

The trained model was then applied to a social media image dataset collected during Hurricane Harvey in 2017. Flood extent was estimated based on this information, which correctly marked over 62\% of the regions where people have claimed flood or water damage. Flood severity was mapped and compared with the modeled flood depth grid. The result indicates a weak positive monotonic correlation to the reference data. In addition, it can serve as a flood severity information which is available well ahead of remote sensing detection.

Regarding future work, we will focus on four aspects. First, in addition to people, other objects with approximately known dimensions can be analyzed to extract water levels, such as vehicles and bicycles. Considering these objects with additional component level information, more flood-related images can be used to increase the number of effective observations. 
Second, since this work mainly focused on image interpretation, text understanding can also be combined into the proposed process, e.g. to improve water extent mapping with flood relevant posts without photos, to parse locations from texts to include more visual observations in the mapping process.
Third, in this work, we analyzed the data collected during Hurricane Harvey offline, however, the efficiency of the entire system needs to be further investigated to test how much time the VGI can be ahead of the information from remote sensing flood detection in practice.
Last, video is another important data generated by social media users and can be included in the framework. In addition, data from surveillance cameras in cities can also be considered, which can provide more observations for flood events, with well-known geolocation.

\section*{Acknowledgement}
The authors would like to acknowledge the support from the BMBF funded research project ``TransMiT -- Resource-optimized transformation of combined and separate drainage systems in existing quarters with high population pressure'' (BMBF, 033W105A) and ``EVUS -- Real-Time Prediction of Pluvial Floods and Induced Water Contamination in Urban Areas'' (BMBF, 03G0846A). We also gratefully acknowledge the support of NVIDIA Corporation with the donation of a GeForce Titan X GPU used for this research.

\bibliographystyle{cas-model2-names}

% Loading bibliography database
\bibliography{cas-dc-template}

%\vskip3pt

% \bio{}
% Author biography without author photo.
% Author biography. Author biography. Author biography.
% \endbio

% \bio{figs/pic1}
% Author biography with author photo.
% Author biography. Author biography. Author biography.
% \endbio

\end{document}